\documentclass[10pt,twocolumn,letterpaper]{article}

\usepackage[pagenumbers]{cvpr} 

\usepackage{graphicx}
\usepackage{amsmath}
\usepackage{amssymb}
\usepackage{booktabs}
\usepackage{silence}
\usepackage[pagebackref,breaklinks,colorlinks]{hyperref}

\usepackage[capitalize]{cleveref}
\crefname{section}{Sec.}{Secs.}
\Crefname{section}{Section}{Sections}
\Crefname{table}{Table}{Tables}
\crefname{table}{Tab.}{Tabs.}

\def\cvprPaperID{2568} 
\def\confName{CVPR}
\def\confYear{2023}

\begin{document}

\title{HRVQA: A Visual Question Answering Benchmark \\for  High-Resolution Aerial Images}


\author{Kun Li, George Vosselman, Michael Ying Yang\\
University of Twente
}
\maketitle

\begin{abstract}
  Visual question answering (VQA) is an important and challenging multimodal task in computer vision. 
  Recently, a few efforts have been made to bring VQA task to aerial images, due to its potential real-world applications in disaster monitoring, urban planning, and digital earth product generation. However, not only the huge variation in the appearance, scale and orientation of the concepts in aerial images, but also the scarcity of the well-annotated datasets restricts the development of VQA in this domain. In this paper, we introduce a new dataset, HRVQA, which provides collected 53512 aerial images of 1024 $\times$ 1024 pixels and semi-automatically generated 1070240 QA pairs. To benchmark the understanding capability of VQA models for aerial images, we evaluate the relevant methods on HRVQA. Moreover, we propose a novel model, GFTransformer, with gated attention modules and a mutual fusion module. The experiments show that the proposed dataset is quite challenging, especially the specific attribute related questions. Our method achieves superior performance in comparison to the previous state-of-the-art approaches.
  The dataset and the source code will be released at
  \url{https://hrvqa.nl/}.
\end{abstract}

\section{Introduction}
\label{sec:intro}
Multimodal learning has gained broad attention from computer vision community recently. For visual-lingual tasks, such as image captioning \cite{anderson2018vqamethod6, xu2015ic}, visual grounding \cite{yu2018vg}, and visual question answering (VQA) \cite{marino2019okvqa, vqa}, impressive progress has been made with powerful deep learning models. Compared to other multimodal tasks, VQA is a more challenging problem as it requires semantic understanding and detailed reasoning over the visual contents and textual elements. VQA for aerial images is a quite new task that has several potential real-world applications such as land use and disaster monitoring and environmental or urban planning \cite{rahnemoonfar2021floodnet}. However, the recent fine-tuning based methods \cite{lobry2020rsvqa} developed for natural scenes cannot handle the numerous difficulties for aerial images, including scale variation, object density, arbitrary orientation, and complex background. The numerous tiny objects in high-resolution aerial images require more specific descriptions in the questions to avoid ambiguity, and make the learning more difficult without any semantic supervision in VQA.

\begin{figure}[t!]
  \centering
  \begin{subfigure}[t]{0.45\linewidth}
  \includegraphics[width=1\linewidth]{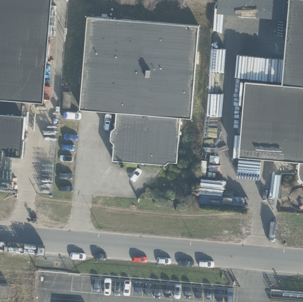}
  \subcaption*{{\textbf{$Q$:}} Are there more buildings than large vehicles in this image?\\ {\textbf{$Q$:}} How many small vehicles are there in this image?}
  \label{fig:example-1}
  \end{subfigure}
  \hfill
  \begin{subfigure}[t]{0.45\linewidth}
  \includegraphics[width=1\linewidth]{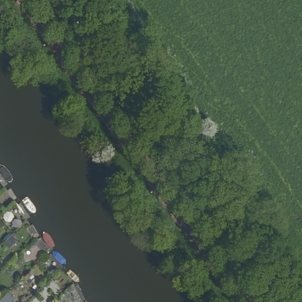}
  \subcaption*{{\textbf{$Q$:}} What color is the second ship based on the left to right rule in this image?\\ {\textbf{$Q$:}} Where is the storage tank in this image?}
  \label{fig:example-2}
  \end{subfigure}
  \caption{Examples of image/question pairs from HRVQA. Ten types of questions are created, covering the inferences from common-seen task reasoning to specific attribute recognition.}
  \vspace{-5mm}
  \label{fig:examples}
\end{figure}

In addition to aforementioned difficulties, current research still lacks comprehensive well-annotated datasets, which are crucial to deep learning models. 
Although there do exist some public datasets of remote sensing images for VQA \cite{lobry2020rsvqa,zheng2021mutual}, they are created with general questions and restricted in terms of spatial resolution, richness of concepts, complexity and diversity of questions, which make it hard to evaluate the real performance of these models. As it is so expensive to manually label the image/question pairs, human efforts need to be taken into account when creating a VQA dataset \cite{yi2019clevrer, zheng2021mutual}. Thus, a smart way of building a well-annotated dataset that contains not only general questions but specific details of the concepts is in high demand.

To advance the VQA research in computer vision and earth observation, we introduce a new dataset, HRVQA (High-Resolution aerial image Visual Question Answering). We collect 53512 aerial images and 1070240 QA pairs in total, which include 10 types of questions with respect to 27 category concepts. Given the quantity, diversity, and novelty of the proposed dataset, HRVQA shall be able to develop as a strong benchmark for aerial image VQA. As shown in Fig.~\ref{fig:examples}, these questions correspond to various scenes involving different concepts in aerial images.

Attention modules embedded in deep learning networks focus on the most relevant contents to improve performance for many computer vision tasks. They have been incorporated in VQA networks and also led to better performances for VQA \cite{dua2021vqamethod3, goyal2017method4, shih2016attentionmethod1}. Co-attention method \cite{kim2018banmethod7} parallelly derives the attended visual and textual features from individual regions and texts. Inspired by the transformers\cite{vaswani2017transformer}, deep modular co-attention networks \cite{yu2019mcan, aoa} address this problem by two units: self-attention (SA) and guided attention (GA). SA and GA extract intra-modal and cross-modal interactions with multi-head attentions, respectively. However, these attention units always calculate the weighted combination of features among the attended modalities, and ignore the positional information of the target concepts. This limitation leads worse results especially when asked a question about an isolated object with specific positional restrictions, which impairs the performance on aerial image VQA dealing with numerous tiny concepts. 

In this paper, we propose GFTransformer, a modular gated co-attention network with a mutual fusion module. Inspired by ConVit \cite{d2021convit}, gated attention modules with positional embedding are employed to extend the previous modular co-attention network \cite{yu2019mcan}. Specifically, the positional embedding is first used to represent the spatial features when fed into a multi-head attention architecture. To make a balance between the content and positional term, a gated transformation is introduced on the combination of the attended features from both sides. By normalizing the gated results, we generate the focused features through multiple attention heads and remove the interference of the irrelevant attentions to achieve reliable reasoning. 
Furthermore, a mutual fusion module is proposed to selectively extract the most useful information from the refined cross-modal features before predicting answers.
Our \textbf{main contributions} of this paper are summarized as follows:
\begin{itemize}
    \item To the best of our knowledge, HRVQA is the first high-resolution aerial image VQA dataset with diverse and concerning questions about not only overall information but specific attributes of various concepts. It can be used to evaluate the capabilities of VQA models on performing scene understanding and geo-spatial reasoning in aerial images.
    \item We propose an efficient labeling scheme for aerial image VQA dataset construction in a semi-automatic fashion, so that we can build a comprehensive well-annotated VQA dataset with a few manual efforts.
    \item In addition to the dataset, we also propose a novel model (GFTransformer) and benchmark several state-of-the-art VQA models, which could serve as the baseline for future research.
\end{itemize}

\section{Related Work}

\noindent
\textbf{Datasets for VQA.} Almost all the deep learning methods, especially strongly supervised algorithms, require a large amount of annotated data for training. Recently, there emerges a great interest in visual and textual context understanding tasks \cite{vqa,imagecaptioning,visualgrounding}, and many VQA datasets \cite{vg,yi2019clevrer,marino2019okvqa} have been introduced in computer vision. Among them, DAQUAR \cite{malinowski2014daquar} is the first VQA dataset designed as a benchmark, which is built with the images of indoor scenes and questions about the basic information of various furniture. VQA 2.0 \cite{goyal2017method4}, one of the most widely used VQA datasets, focuses on both wild-world and indoor scenes for various questions with the images from COCO \cite{lin2014coco}. GQA \cite{hudson2019gqa} presents natural images with scene graphs and questions requiring diverse reasoning skills and sequential steps to generate answers. With respect to the field of earth observation, there have also been some pioneers delving into VQA \cite{rahnemoonfar2021floodnet,yuan2022easy}. RSVQA \cite{lobry2020rsvqa} consists of one low-resolution subset and one high-resolution subset with simple QA pairs. RSIVQA \cite{zheng2021mutual} is composed of available image classification and object detection datasets with restricted QA pairs generated based on the annotations. Differently, our HRVQA focuses on the intelligent perception and reasoning in high-resolution aerial images with more diverse questions.

\vspace{2mm}
\noindent
\textbf{QA Pair Generation.} Question Generation (QG) is a quite active research field in natural language processing \cite{du2017learningqg1,lopez2020transformerqg2}, which could be used as a preparation stage for other textual tasks, including question answering \cite{duan2017qgqa1} or information retrieval \cite{mass2020unsuperviseqgir}. In computer vision, it aims at generating QA pairs for a given image (or video) as training data or means of data augmentation for VQA \cite{li2018vg4qaonly333}. In this paper, our QA pair generation is based on a template strategy that requires the corresponding collection of visual information to answer the generated questions.

\vspace{2mm}
\noindent
\textbf{VQA Methods.} 
In the last decade, we have witnessed the rising and fast pace developments of deep learning models for VQA \cite{goyal2017method4,wang2015vqamethod1,kim2018banmethod7,zhou2020method13,yu2019mcan}. In the general framework for VQA, a model needs to process the input image and question as global features separately and then fuse them together with a multimodal fusion module to predict the answers. RSVQA \cite{lobry2020rsvqa} included a feature extraction part based on ResNet-152 \cite{he2016deepresnet} and skip-thought \cite{kiros2015skip}, a point-wise multiplication layer for fusion and a classifier for prediction. However, only global features may mislead the learning process for some questions requiring local information. Attention techniques can handle the shortcoming by focusing on the most relevant visual regions or textual phases. SAN \cite{yang2016stackedsan} proposed a stacked attention network to learn the relationships among targets, which helps to discard the irrelevant objects in the image. MCAN \cite{yu2019mcan} adopted a dense co-attention model to learn the textual attention for questions and the visual attention for images simultaneously. Martins et al. \cite{martins2020sparse} and Zhou et al. \cite{zhou2021trar} followed \cite{yu2019mcan} and exploited different compact region selection methods to extract valid features. Whitehead et al. \cite{whitehead2022reliableeccv2022} utilized a multimodal selection function to estimate the correctness of the answers predicted by a co-attention model. One limitation of the aforementioned co-attention models is that the equally treated feature representations extracted from self-attention modules are directly projected into the answer space without further differentiation. Differently, our GFTransformer investigates the weighted information for each query in both modalities to improve the co-attention architecture, and selects the attended features for predictions.
\section{Construction of HRVQA Dataset}
In general, HRVQA dataset contains 1070240 QA pairs and 53512 images from high-resolution aerial scenes. In this section, we first introduce the resources of images and then explain the process of question and answer generation, respectively. Finally, we analyze the visual and lingual properties of the proposed dataset.

\subsection{Image Collection}
To build a VQA dataset with aerial images, there are several unique visual features to be taken into account. For aerial images, the (spatial) resolution is an extremely important property for many tasks, which indicates how detailed the ground contents can be observed. Besides, the images in various scenes need to contain diverse concepts. The aerial images of nadir view need to show the ground targets clearly.

Given that the current public aerial image VQA datasets \cite{lobry2020rsvqa,rahnemoonfar2021floodnet,zheng2021mutual} cannot meet the aforementioned requirements, we collect high-resolution aerial images from Dutch OpenData\cite{quickortho} for our VQA visual part. It is a collaboration of Dutch governmental organizations with the aim of providing nationwide aerial images in a ground pixel resolution of 8 cm. After selecting 4 typical cities in Netherlands, \textit{Utrecht, Rotterdam, Enschede and Amsterdam}, we obtain 53512 nadir-viewed aerial images of $1024 \times 1024$ pixels that contain rich and different concepts in various scenes (e.g., industry, harbor, station, sports).
\subsection{Question Generation}
\label{qg}
As it is tedious to completely include all the questions that one could ask for a given aerial image, our question set construction starts with the most related and common ones. To generate richer and more diverse questions, we consider the construction from three aspects: category concepts, question types and templates. We randomly generate 200 questions for each image to avoid missing of the questions about relatively rare concepts (e.g., plane).

\noindent
\textbf{Category Concepts.} We choose 27 category concepts~\footnote{The examples for these concepts are provided in the supplementary.} (including isolated objects and region stuff), \textit{small-vehicle, large-vehicle, ship, plane, storage tank, baseball diamond, tennis court, basketball court, bridge, helipad, helicopter, soccer ball field, swimming pool, roundabout, ground track field, harbor, airport, track, outdoor fitness field, high-density urban area, low-density urban area, rural area, industrial area, station, sports area, water section, building}. These concepts are selected according to the frequency of occurrence and the usage in our daily activities, which support VQA system to serve for more potential real-world applications. We further introduce two common attributes, \textit{size} and \textit{location}, for every concept, and two special attributes, \textit{shape} and \textit{color}, only for isolated objects. Each attribute is an optional restriction on the concepts to make the reasoning more detailed. 

\noindent
\textbf{Question Types.} The aim of a question lies on the information that a questioner wants to know from the image. Different from the simple classification of question types for natural images, \textit{Yes/No, Number} and \textit{Others}, we further classify the questions to be generated into 10 types according to their functions, \textit{Number, Yes/No, Areas, Location, Color, Shape, Size, Scene, Transportation} and \textit{Sports}. They are closely related to the selected concepts and able to describe various aspects of them (from general to specific information). Note that it is extremely vital to avoid ambiguity for some question types in VQA due to the fact that there is always more than one object with the same category in aerial images. To improve the clarity of the questions, we add a set of more specific descriptions on the related concepts according to the relevant spatial location. For instance, we ask \textit{What color is the second ship in the image based on the left to right rule?} rather than \textit{What color is the ship in the image?} This solves the problem that the questions about isolated objects cannot be asked in the current remote sensing image VQA datasets \cite{lobry2020rsvqa,zheng2021mutual}.

\noindent
\textbf{Question Templates.} Given the randomly selected concept and question type from the corresponding candidate pools, respectively, the main question structure starting with \textit{How, What, Where, Is, Are} and \textit{Can} is defined. Then a relevant template for each question type is randomly selected from a pre-defined template set. For instance, we could ask \textit{How many ships are there in this image?} or \textit{What is the number of ships in this image?} for a \textit{Number} question.
\subsection{Answer Generation}
To obtain the answers to the generated questions, we collect semantics of the involved concepts from images based on an object detection model \cite{yolov5} and the available annotations from Dutch open resource PDOK \cite{pdok}, and then match them with the functions of the proposed questions.

Specifically, DOTA \cite{dota, dota-v2} contains a small set of aerial images as well as bounding box annotations for 18 category objects, which cover the majority of the concepts in our datasets. Given the advances of object detection models pre-trained on large-scale datasets, we utilize a detection model YOLO v-5 \cite{yolov5} and finetune it on the subset to predict the objects in our aerial images. Achieved 78.7\% mAP detection results could hugely alleviate the burdens of manual editing (e.g., tiny and frequent objects like \textit{vehicles, ships}). PDOK \cite{pdok} provides open datasets as well as recent metadata in the form of point lists for each ground target of several categories. Then we extract the spatial information about \textit{tracks, buildings} and \textit{water sections}. Upon the collection of all the semantics is completed, we generate the answers to the questions based on the spatial and category information. Note that we set a flag $\in \{0, 1\}$ for every triplet to indicate its relevance. There are two steps for calculating the relevance: first checking if the image contains the target concept involved in the question, then judging if the attribute of the concept in the image is consistent with the contents in the question.

\begin{figure*}[ht]
  \centering
  \begin{subfigure}[t]{0.35\linewidth}
    \includegraphics[width=1\linewidth]{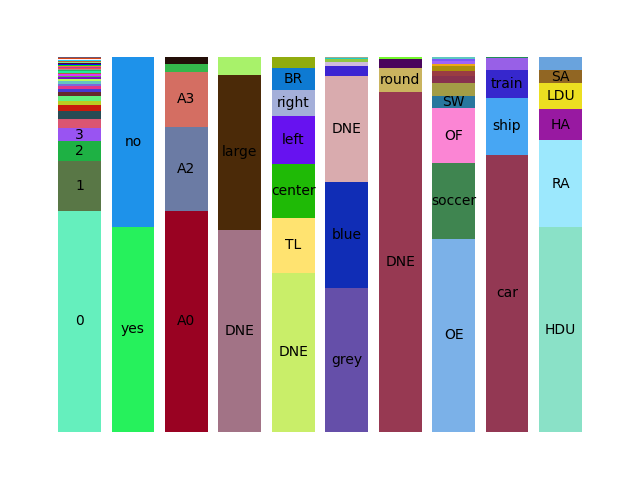}
    \caption{Distribution of Answers Per Question Type.}
    \label{fig:short-a}
  \end{subfigure}
  \hfill
  \begin{subfigure}[t]{0.3\linewidth}
    \includegraphics[width=1\linewidth]{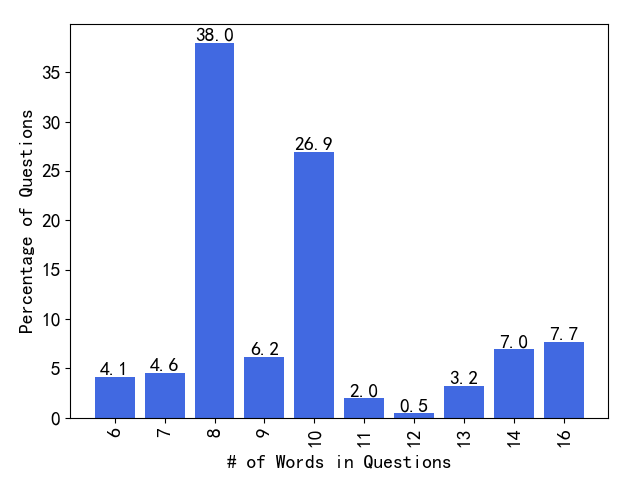}
    \caption{Distribution of Question Lengths.}
    \label{fig:short-b}
  \end{subfigure}
  \hfill
  \begin{subfigure}[t]{0.3\linewidth}
    \includegraphics[width=1\linewidth]{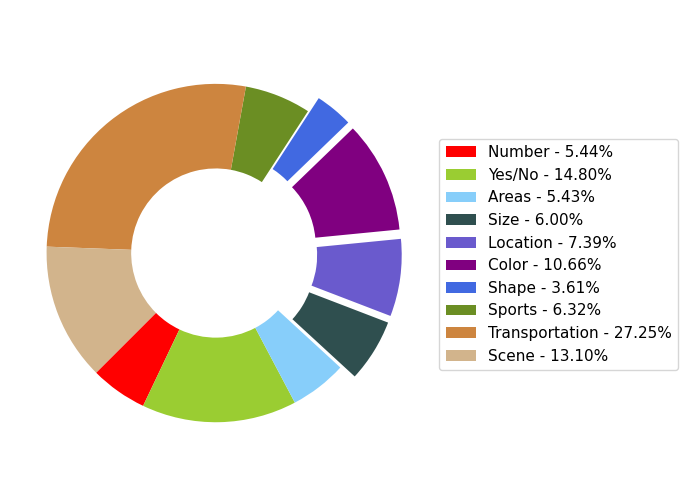}
    \caption{Distribution of Question Types.}
    \label{fig:short-c}
  \end{subfigure}
  
  \caption{Statistics of question/answer pairs in HRVQA. \textbf{Left} represents distribution of answers per question type in the following order: \textit{Number, Yes/No, Areas, Size, Location, Color, Shape, Sports, Transportation and Scene}. The short names for the answers are defined as: \textit{A0-$0m^2$, A2-$between\ 10\ and\ 100m^2$, A3-$between\ 100\ and\ 1000m^2$, DNE-does not exist, TL-top left, BR-bottom left, OE-other exercise, OF-outdoor fitness, SW-swimming, HDU-high-density urban area, RA-rural area, HA-harbor, LDU-low-density urban area, and SA-sports area}.}
  \label{fig:short}
\end{figure*}

We also apply some transformations on the numerical answers inspired by \cite{lobry2020rsvqa}. Differently, only answers to \textit{Areas} question are quantized into the following categories: a) $0m^2$; b) $between\ 0\ and\ 10m^2$; c) $between\ 10\ and\ 100m^2$; d) $between\ 100\ and\ 1000m^2$; e) $more\ than\ 1000m^2$. We insist to predict exact values for the \textit{Number} questions because they are more meaningful and challenging for VQA.
\subsection{Triplet Filter}
\label{triplet filter}
To maintain the balance and diversity of the proposed dataset, we further adjust the distribution of different types of questions and remove the majority of the triplets with irrelevant answers, which mean the meaningless short-cuts whose flags equal 0 in the previous settings. The first step is to reduce the proportion of meaningless triplets. Specifically, among the randomly generated 200 triplets, we reassign 20 QA pairs with 70\% active flags for each image (if not satisfied, repeat the generation scheme until 70\%). Then we replace the half of the questions from transportation and scene types (easy to predict) with new randomly generated questions from other types (hard to predict).

Till this end, a balanced and diverse aerial image VQA dataset has been created. The proposed method for QA pair generation enables the semi-automatic and scalable VQA triplet construction with only a few manual efforts.

\subsection{Properties of HRVQA Dataset}
The images in HRVQA are all from the same airborne source provided by \cite{quickortho}. The collection of the images took place in the leafless season (from mid-February to mid-April), so that as much as possible information can be captured under the trees and vegetation from the images. We collect the RGB-channel aerial images in 8 cm resolution from 4 typical Dutch cities. Table \ref{tab:statistics of images} shows the statistics of images and concepts in terms of the number and covering area. Note that we do not present the covering area for the chosen concepts due to the uncertainty of region stuff’s boundary.

\begin{table}
  \caption{Statistics for images and concepts in HRVQA.}
  \centering
  \setlength\tabcolsep{2.5pt}
  \begin{tabular}{p{1.85cm}cccc}
    \toprule
    Number & Utrecht & Rotterdam & Enschede & Amsterdam\\
    \midrule
    \# of images & 12320 & 14716 & 6762 & 19714\\
    \# of concepts & 141033 & 206306 & 71705 & 236152\\
    Area ($km^2$) & 100 & 110 & 72 & 156\\
    \bottomrule
  \end{tabular}
  \label{tab:statistics of images}
  \vspace{-5mm}
\end{table}
Moreover, we show the statistics of the language distribution in the proposed dataset. In general, HRVQA covers 10 interesting and meaningful types of questions requiring scene understanding and reasoning inference. Fig. \ref{fig:short-a} presents the distribution of answers per question type. We can see that some question types, such as \textit{Size, Shape}, and \textit{Transportation} are dominated by one or two answers due to the real circumstance under the bird's view (e.g., roundabouts are always round; car is the most common transportation in the cities). Other question types are more even in answers. We further show the distribution of question lengths counted by every single word in Fig. \ref{fig:short-b}, and $9.7$ is the average value for our dataset. Fig. \ref{fig:short-c} indicates the distribution of question types, which indirectly implies the distribution of different questions about general and detailed information.
\section{Method}
\label{method}
\begin{figure*}[ht]
  \centering
  \includegraphics[width=0.85\linewidth]{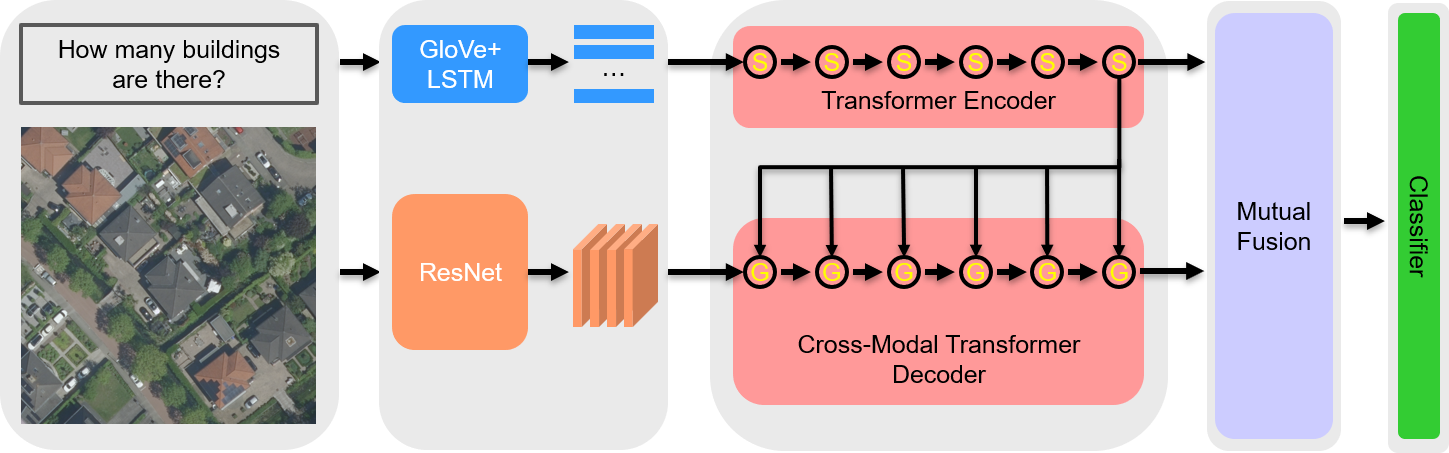}
  \caption{Pipeline of the proposed GFTransformer. The different units in the Encoder-Decoder layers represent self multi-head attentions (S) and guided multi-head attentions (G) with gated positional embeddings. With the Mutual Fusion module, the attended features are selectively combined to predict a desired answer.}
  \label{fig:pipeline}
  \vspace{-5mm}
\end{figure*}

To deal with high-resolution aerial image VQA, we propose GFTransformer (Gated attention and mutual Fusion Transformer), as illustrated in Fig.~\ref{fig:pipeline}. We first introduce the input feature representations from images and questions, and explain self-attention and guided-attention units adopted in our baseline model \cite{yu2019mcan}. Then we propose a gated attention module to extract the attended features with positional information. By cascading a number of aforementioned layers in depth, an encoder-decoder model refines the attended intra-modal and cross-modal features. Finally, a mutual fusion module is proposed to selectively combine the attended visual and lingual features together before projected into answer space.

\vspace{2mm}
\noindent
\textbf{Input Feature Representations.} Given that VQA models develop cross-modal mathematical operations on multi-source data, the images and questions need to be represented as vectors by using different feature extraction methods, respectively. \cite{anderson2018vqamethod6} first represents the images as a set of regional features in a bottom-up and top-down mode, which is powerful for the natural scenes and thus adopted in many VQA models \cite{anderson2018vqamethod6,zhou2020method13,zhou2021trar}. In contrast to natural images, aerial images cover numerous tiny objects in some cases (e.g., vehicles in a parking lot), which impair the capability of regional representations. To overcome this limitation, we replace the regional visual features in our baseline model \cite{yu2019mcan} with grid features $X$ generated by ResNet-152 \cite{he2016deepresnet}. Following \cite{yu2019mcan}, questions are represented into 512-D vectors by the GloVe word embeddings \cite{pennington2014glove} followed a LSTM \cite{hochreiter1997lstm} layer, to generate the question features $Y$.

\vspace{2mm}
\noindent
\textbf{Transformer Units.} Multi-head attention layers parallelly develop $N$ heads to capture different focuses on the features, where each head could be conducted by a scaled dot product attention function. The standard transformer attention function is shown as follows:
\begin{equation}
    f_{trans} = A(Q,K,V) = {\rm Softmax}(\frac{QK^T}{\sqrt{d}})V,
\end{equation}
where $Q,K,V$ indicate the queries, keys, and values, respectively, and $d$ represents the dimension of keys and values (normally set the same number for both). The function outputs the weighted summation over the values with attended information obtained from keys and queries. Specifically, the results of dot products between queries and keys are divided by $\sqrt{d}$ as attentions, and then the attentions are normalized to generate new weighted values.

Self-attention unit (SA) and guided-attention unit (GA) are built on the top of the multi-head attentions followed a feed-forward network (FFN) with ReLU activation and dropout. The residual connection \cite{he2016deepresnet} is adopted to maintain the reliable information and address gradient issues. The main difference between these two units is that SA only takes the input features from one modality for the intra-modal learning (word-to-word and region-to-region) while GA learns to capture the cross-modal relationships (word-to-region, region-to-word). Specifically, GA takes the queries from one modality to guide the attention learning based on the keys and values from the other one while SA only handles the individual elements.

\begin{figure}[t]
  \centering
  \includegraphics[width=0.57\linewidth]{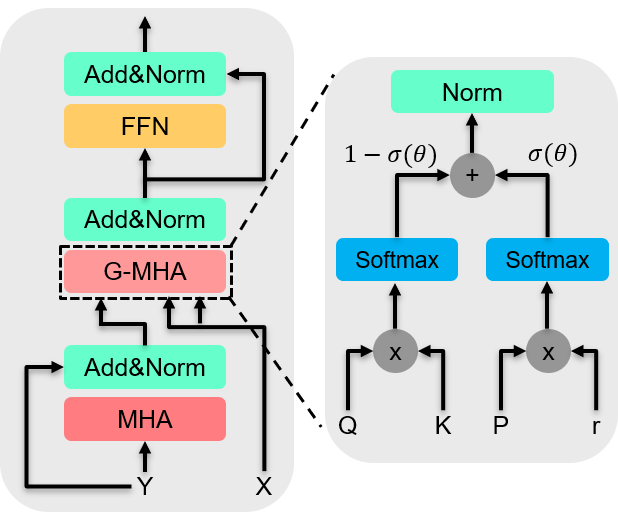}
  \caption{Gated attention unit with multi-head attention (MHA). \textit{X} and \textit{Y} denote the visual and lingual features, respectively.}
  \label{fig:gated attention}
  \vspace{-5mm}
\end{figure}

\vspace{2mm}
\noindent
\textbf{Gated Attention Module.} The aforementioned attention units capture the relations between $Q$ and $K,V$, and output the weighted features without discriminate selection, which may mislead the answering process. To overcome the limitation, we add a gated attention module based on a positional guidance. In detail, the positional encoding $r_{xy} $ is used to represent the relative position (distance representations) of patches $x$ and $y$ among grid regions, and the corresponding embedding $P$ for this relation could be further trained with attention heads. We fix the relative positional encodings and make the network only learn the embedding to discover the adaptive attention span \cite{sukhbaatar2019adaptivespan}. The learned attention is shown as follows:
\begin{equation}
    f_{atte} = {\rm Softmax}(\frac{QK^T}{\sqrt{d}}+P^Tr_{xy}),
\end{equation}
As the magnitudes of both terms always face the issue of mismatching, ${\rm Softmax}$ cannot take the content and positional information into account perfectly at the same time. To alleviate the instability of the learning process, we adopt a learnable parameter $\theta$ in each attention head as the factor to balance the combination of both sides, which is a useful strategy to measure the importance of different terms \cite{d2021convit}. With normalization on the summation of both terms, the gated positional attention is completed as follows:
\begin{equation}
    f_{atte} = {\rm Normalize}((1-\sigma(\theta))f_c+\sigma(\theta)f_p),
\end{equation}
\begin{equation}
    f_c = {\rm Softmax}(\frac{QK^T}{\sqrt{d}}),   f_p = {\rm Softmax}(P^Tr_{xy}),
\end{equation}
where $\sigma$ is a sigmoid function. We initialize $\theta=1$ for all the heads and layers in the transformer architecture. Fig. \ref{fig:gated attention} shows the architecture of the gated attention unit.

Till this end, we feed the visual features $X$ and question features $Y$ into an encoder-decoder transformer with cascaded gated attention units. Simply, the output of the last attention layer is used as the input for the next one. We refine the features by stacking 6 gated attention units for intra-modal and cross-modal leaning in our encoder-decoder transformer~\footnote{More details about the architecture can be found in the supplementary.}. 

\begin{figure}[t]
  \centering
  \includegraphics[width=0.8\linewidth]{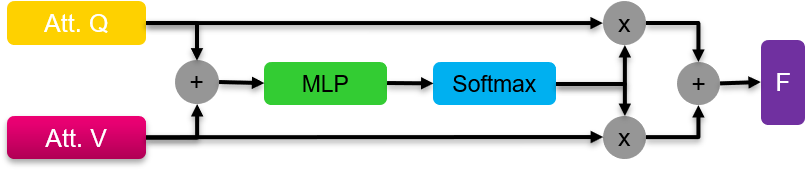}
  \caption{Mutual fusion module after the Encoder-Decoder. Summation operation followed a \textit{MLP} and a ${\rm Softmax}$ helps to select the attended features from both modalities, to generate fusion results \textit{F}.}
  \label{fig:mutual fusion}
  \vspace{-5mm}
\end{figure}

\vspace{2mm}
\noindent
\textbf{Mutual Fusion Module.} 
Yu et al. \cite{yu2019mcan} uses a simple multimodal fusion model based on a two-layer MLP (fully connected layer, ReLU, and dropout) to fuse the individual attended features, which is modal-agnostic. In this paper, we propose a mutual fusion module to project the features from both modalities into a better output space. First, the same MLP is adopted to generated attended features $X'$ and $Y'$, respectively. Different from \cite{yu2019mcan}, we use the features in one modality to guide the refining of the features from the other. Specifically, we summarize $X'$ and $Y'$ followed by a three-layer MLP (fully connected layer and dropout). To determine the impact of information from each modality, we add a ${\rm Softmax}$ on the summation and generate new mutual weighted features $X''$ and $Y''$ with the similar operations for $X'$ and $Y'$, respectively. With two linear projection matrices on $X''$ and $Y''$ followed a ${\rm LayerNorm}$ layer for stable training, we obtain the fusion results for the final predictions, which is realized by a sigmoid activation. Fig. \ref{fig:mutual fusion} illustrates the architecture of the proposed mutual fusion module.

\section{Experiments}
In this section, we conduct experiments to verify the effectiveness of our GFTransformer on the proposed HRVQA dataset. We first present the setup for the experiments, and then perform a benchmark based on several state-of-the-art VQA models. To further analyze the effectiveness of core model designs, we conduct extended ablation studies. Finally, comparative experiments on RSVQA-HR \cite{lobry2020rsvqa} are completed to validate the robustness of our model.

\begin{table*}
  \caption{Benchmark results of type-wise accuracy on HRVQA. The short names for the question types are defined as: $Y/N-Yes/No, Locat.-Location, Trans.-Transportation$. Note that $\dagger$, $\ddagger$, and $\ast$ represent basic CNN models, attention based CNN models, and transformers, respectively.}
  \centering
  \setlength\tabcolsep{5.5pt}
  \begin{tabular}{p{2cm}cccccccccccc}
    \toprule
    Method & Number & Y/N & Areas & Size & Locat. & Color & Shape & Sports & Trans. & Scene & OA & AA \\
    \midrule
    SAN\cite{yang2016stackedsan}$\ddagger$   & 62.89 & 82.15 & 86.30 & 93.48 & 60.19 & 61.97 & 92.13 & 72.86 & 87.72 & 74.84 & 79.01 & 77.45 \\
    MUTAN\cite{ben2017mutanmethod10}$\ddagger$  & 62.05 & 87.44 & 83.43 & 93.41 & 59.75 & \textbf{70.66} & 87.58 & 61.59 & 82.51 & 73.96 & 77.97 & 76.24 \\
    MCAN\cite{yu2019mcan}$\ast$   & 66.53 & 93.17 & 97.88 & 93.27 & 60.52 & 44.04 & 96.44 & \textbf{78.31} & 88.71 & 77.68 & 80.69 & 79.66 \\
    RSVQA\cite{lobry2020rsvqa}$\dagger$ & 61.97 & 84.81 & 75.50 & 93.41 & 63.96 & 67.85 & 87.05 & 62.34 & 81.82 & 67.98 & 76.19 & 74.67 \\
    M-CA\cite{martins2020sparse}$\ast$   & 66.25 & 93.22 & 97.93 & 93.30 & 61.23 & 44.48 & 96.46 & 77.28 & 88.71 & 77.92 & 80.75 & 79.68 \\
    TRAR\cite{zhou2021trar}$\ast$   & 66.86 & 93.31 & 97.26 & 93.47 & 63.20 & 44.50 & 96.51 & 76.76 & 88.84 & \textbf{78.69} & 81.02 & 79.94\\
    AOA\cite{aoa}$\ast$ & \textbf{66.97} & \textbf{93.41} & \textbf{97.98} & 93.55 & 62.20 & 44.83 & 96.37 & 77.70 & 88.68 & 78.12 & 80.98 & 79.98 \\
    FETH\cite{yuan2022easy}$\dagger$ & 62.15 & 87.86 & 82.37 & 93.44 & 62.88 & 68.13 & 90.59 & 72.93 & 84.23 & 77.07 & 79.49 & 78.17 \\
    Swap-m\cite{gupta2022swapmix}$\ast$ & 63.98 & 89.60 & 94.56 & 90.05 & 59.07 & 43.42 & 92.81 & 76.34 & 85.28 & 77.61 & 78.26 & 77.27 \\
    GFT (ours)$\ast$   & 66.50 & 93.32 & 97.11 & \textbf{93.72} & \textbf{74.82} & 45.36 & \textbf{96.67} & 77.03 & \textbf{88.87} & 77.30 & \textbf{81.71} & \textbf{81.07}\\
    \bottomrule
  \end{tabular}
  \label{tab:results in vhrqa}
  \vspace{-5mm}
\end{table*}
\subsection{Experiment Setup}
\label{setup}
\noindent
\textbf{Datasets.} To evaluate the proposed method, the main experiments are conducted on the proposed high-resolution aerial image dataset, HRVQA. To fairly compare the performances of different models designed for natural scenes with ours, we re-trained these models on HRVQA. We split the dataset into a training set (\textit{Utrecht} and \textit{Rotterdam}), a validation set (\textit{Enschede}), and a test set (\textit{Amsterdam}), and their proportions are 50.5\%, 12.6\%, and 36.9\%, respectively. To further verify the generalization and robustness of the proposed method, we conducted extensive experiments on another aerial image dataset RSVQA-HR\cite{lobry2020rsvqa} that contains a considerable number of image/question/answer triplets.

\vspace{2mm}
\noindent
\textbf{Implementation Details.} To make a fair comparison, we conducted all the experiments on the same \textit{PyTorch} environment with 1 NVIDIA RTX 2080 Ti GPU in 11GB memory, and trained the models by using the \textit{Adam} optimizer with $\beta_1=0.9$ and $\beta_2=0.98$ for 13 epochs. We set the base learning rate to the minimum of 0.0000025 and 0.0001 and decay it by 0.2 at epoch 10 and 12. The batch size is set to 32, and the latent dimension in the multi-head units, the number of heads, and the latent dimension for each head are set to 512, 8, 64, respectively. More details can be found in the supplementary material. We used the mainstream overall accuracy (OA) and average accuracy (AA) for the evaluation statistics.

\subsection{Benchmark Results}
To promote the development of VQA in earth observation, we built a benchmark by conducting experiments on the proposed HRVQA over different methods, including basic CNN models, attention based CNN models, and transformer models. Besides, we provided a new baseline model for aerial image VQA, and compared the results with others. The comparative results on the test set are listed in Table \ref{tab:results in vhrqa}. We observed the following benchmark results: (1) Our proposed method achieves the best overall accuracy (81.71) and average accuracy (81.07) on HRVQA. (2) Different methods are proposed for various directions in VQA, based on their specially designed modules, training schemes and so on. This leads their performances differ in individual question types. For instance, AOA \cite{aoa} handles the common-seen question better than the others (e.g., \textit{Number} and \textit{Yes/No} are classical questions for natural images), while our method performs better in the positioning related questions (e.g., \textit{Size, Location, Shape}). (3) CNN models \cite{lobry2020rsvqa, yuan2022easy, yang2016stackedsan, ben2017mutanmethod10} incline to perform better than transformers \cite{aoa,yu2019mcan, zhou2021trar} in the term of \textit{Color} questions, which is nearly 25\% gap. On the other hand, these questions are always about the information of isolated objects (e.g., \textit{small vehicle, large vehicle, ship}), which are significantly smaller than the others. We conjecture that convolution and pooling operations are more effective in capturing the color attribute of tiny objects from aerial images when the supervision of corresponding semantics is absent in VQA.

\begin{figure*}[ht]
  \centering
  \begin{subfigure}[t]{0.45\linewidth}
    \includegraphics[width=1\linewidth]{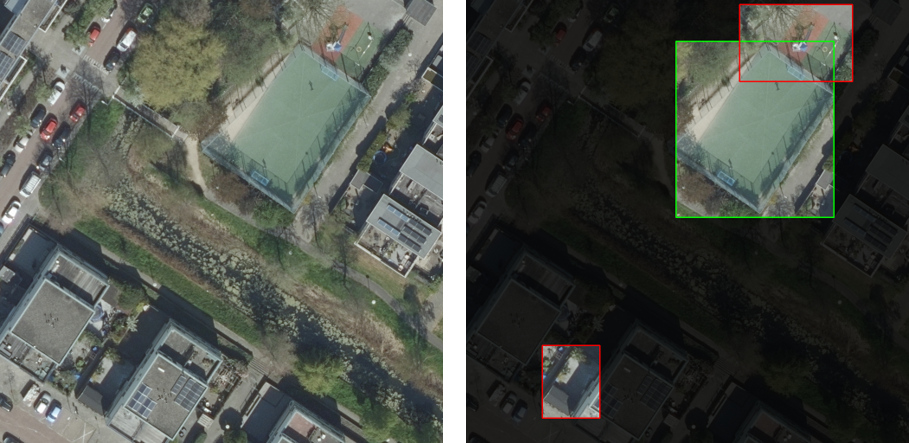}
    \subcaption*{\textbf{$Q$:} What sport can people do in this image? \textbf{$A$:} playing soccer}
  \end{subfigure}
  \hspace{8mm}
  \begin{subfigure}[t]{0.45\linewidth}
    \includegraphics[width=1\linewidth]{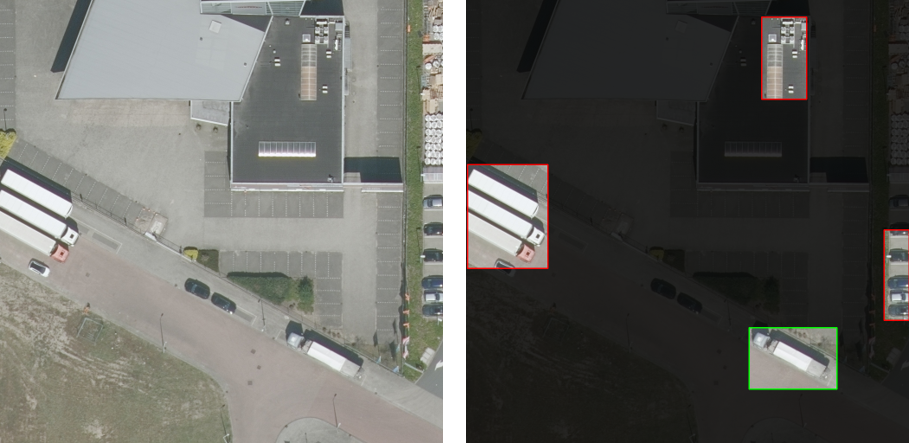}
    \subcaption*{\textbf{$Q$:} What is the color of the third large vehicle based on the left to right rule in this image? \textbf{$A$:} grey}
  \end{subfigure}
  \vspace{-2mm}
  \caption{Visualizations of results from GFTransformer on HRVQA. \textbf{Right} side of each pair shows the attentions across bounding boxes, and the region in a green box means the highest weight included.}
  \label{fig:visualization}
  \vspace{-2mm}
\end{figure*}

\begin{table*}
  \caption{Ablation studies for individual modules. G-MHA and MF represent gated attention unit and mutual fusion module, respectively.}
  \centering
  \setlength\tabcolsep{5.5pt}
  \begin{tabular}{p{2.2cm}cccccccccccc}
    \toprule
    Method & Number & Y/N & Areas & Size & Locat. & Color & Shape & Sport & Trans. & Scene & OA & AA \\
    \midrule
    Baseline\cite{yu2019mcan}   & 66.53 & 93.17 & 97.88 & 93.27 & 60.52 & 44.04 & 96.44 & 78.31 & 88.71 & 77.68 & 80.69 & 79.66\\
    +\textit{only} G-MHA & 66.57 & 93.24 & 97.55 & 92.89 & 74.15 & 44.72 & 96.63 & 76.59 & 88.15 & 77.66 & 81.37 & 80.81 \\
    +\textit{only} MF & \textbf{66.84} & \textbf{94.34} & \textbf{98.22} & 93.40 & 61.12 & \textbf{45.76} & 96.39 & \textbf{78.44} & 88.87 & \textbf{78.03} & 81.07 & 80.04\\
    GFT (ours)   & 66.50 & 93.32 & 97.11 & \textbf{93.72} & \textbf{74.82} & 45.36 & \textbf{96.67} & 77.03 & \textbf{88.87} & 77.30 & \textbf{81.71} & \textbf{81.07}\\
    \bottomrule
  \end{tabular}
  \label{tab:ablation1 in vhrqa}
  \vspace{-2mm}
\end{table*}


In summary, the proposed benchmark comprehensively introduces a challenging problem about VQA for aerial images. It contains not only the common questions but also the new designed questions for aerial images. Moreover, the proposed method outperforms the state-of-the-art models in overall accuracy and average accuracy, respectively. Fig. \ref{fig:visualization} shows some examples of our predictions and attention across bounding boxes. 


\subsection{Ablation Studies}
\label{ablation}
To verify the effectiveness of the proposed individual modules and evaluate different input representations, we design a number of ablation studies. More information can be found in the supplemental material.

\vspace{2mm}
\noindent
\textbf{Effectiveness of Proposed Modules.} In our proposed GFTransformer, there are two core improvements over the baseline model: gated attention unit (G-MHA) and mutual fusion module (MF). G-MHA is present to add positional attentions on the features while MF tries to combine the attended features from both modalities together. Table \ref{tab:ablation1 in vhrqa} shows the results of individual module added on the baseline model \cite{yu2019mcan}. When only developing MF, the overall accuracy is improved by 0.38\%. And 0.68\% improvement is achieved with only G-MHA. Moreover, when combining the two components together, we get the best overall accuracy 81.71\% and average accuracy 81.07\%, which indicates that they could complement each other.

\vspace{2mm}
\noindent
\textbf{Different Input Feature Representations.} To figure out the performances of different feature representations, we further ran four groups of combinations: region features + GloVe features \cite{pennington2014glove}, region features + Bert (a pre-trained model in machine translation) features \cite{devlin2018bert}, grid features + GloVe features \cite{pennington2014glove}, and grid features + Bert features \cite{devlin2018bert}. The results are listed in Table \ref{tab:ablation3}. Grid features are more suitable for aerial image VQA due to the resolution and the size of concepts. Pre-trained Bert \cite{devlin2018bert} does not improve the results when dealing with the questions from HRVQA.

\begin{table}
  \caption{Ablation studies for different input feature representations on HRVQA.}
  \centering
  \setlength\tabcolsep{12pt}
  \begin{tabular}{p{3cm}cc}
    \toprule
    Method & OA & AA \\
    \midrule
    Region + GloVe\cite{pennington2014glove}  & 80.78 & 79.25 \\
    Region + Bert\cite{devlin2018bert}  & 80.73 & 79.43 \\
    Grid + Bert\cite{devlin2018bert}  & 81.25 & 80.63 \\
    Grid + GloVe\cite{pennington2014glove}    & \textbf{81.71} & \textbf{81.07} \\
    \bottomrule
  \end{tabular}
  \label{tab:ablation3}
  \vspace{-5mm}
\end{table}
\subsection{Results on RSVQA}
To further evaluate the robustness of the proposed method on various datasets, we trained and tested different VQA models on the high-resolution subset (RSVQA-HR \cite{lobry2020rsvqa}) of the available aerial image dataset RSVQA. RSVQA-HR has two test sets called Test-1 and Test-Phila. Test-1 contains the similar regions as the training samples while Test-Phila covers the regions from another city called Philadephia, which are not seen during the training process. Table \ref{tab:results in rsvqa test phili} shows the comparative experimental results on RSVQA-HR Test-Phila.
We could see the significant improvements with our proposed GFTransformer. Specifically, compared to the state-of-the-art VQA model designed for aerial images \cite{yuan2022easy}, 2.45\% overall accuracy and 2.70\% average accuracy boosts are obtained on Test-Phila. Besides, we perform better than the models designed for natural scenes \cite{aoa,ben2017mutanmethod10,yang2016stackedsan,zhou2021trar,martins2020sparse} in general with 1.53\% overall accuracy and 1.74\% average accuracy improvements. 

\begin{table}
    \caption{Results of type-wise accuracy on RSVQA~\cite{lobry2020rsvqa} Test-Phila.}
  \centering
  \setlength\tabcolsep{2.75pt}
  \begin{tabular}{p{2cm}cccccc}
    \toprule
    Method & Pres & Count & Comp & Area & OA & AA \\
      & -ence &  & -arison &  &  &  \\
    \midrule
    SAN\cite{yang2016stackedsan}$\ddagger$  & 87.95 & 61.24 & 87.83 & 77.98 & 79.48 & 78.75 \\
    MUTAN\cite{ben2017mutanmethod10}$\ddagger$ & 87.50 & \textbf{62.47} & 88.39 & 79.75 & 80.13 & 79.53 \\
    MCAN\cite{yu2019mcan}$\ast$   & 87.73 & 62.28 & 86.41 & 83.12 & 79.97 & 79.88 \\
    RSVQA\cite{lobry2020rsvqa}$\dagger$ & 86.26 & 61.47 & 85.94 & 76.33 & 78.23 & 77.50 \\
    M-CA\cite{martins2020sparse}$\ast$   & 87.54 & 61.63 & 86.94 & 84.64 & 80.15 & 80.19 \\
    TRAR\cite{zhou2021trar}$\ast$   & 87.11 & 61.92 & 87.27 & 86.28 & 80.47 & 80.65 \\
    AOA\cite{aoa}$\ast$   & \textbf{89.08} & 61.26 & 88.27 & 86.04 & 81.11 & 81.17 \\
    FETH\cite{yuan2022easy}$\dagger$ & 87.97 & 61.95 & 87.68 & 78.62 & 79.29 & 79.06 \\
    GFT (ours)$\ast$   & 89.00 & 61.71 & \textbf{89.56} & \textbf{86.77} & \textbf{81.74} & \textbf{81.76} \\
    \bottomrule
  \end{tabular}
  \label{tab:results in rsvqa test phili}
  \vspace{-5mm}
\end{table}

\section{Conclusion}
In this paper, we propose a new VQA dataset for high-resolution aerial images, HRVQA, to build a comprehensive benchmark for computer vision and earth observation research. We introduce a semi-automatic construction scheme for labeling the image/question pairs with 10 concerning types of questions. Moreover, we provide a novel network GFTransformer for aerial image VQA, which could serve as a baseline for future research. The experiments show that our method achieves the best results for aerial image VQA.

\appendix
\section*{Appendix}
In this supplementary material, we provide more details about the experiments on the proposed dataset HRVQA. First, we further show more implementation details about the experiments in Sec.\ \ref{implementation details}. Then in Sec.\ \ref{dataset construction}, we describe the dataset construction process for our high-resolution aerial image VQA dataset, which  includes language settings, semantic extraction, triplet construction, and meaningless triplet filtering. After that, in Sec.\ \ref{more ablation studies} we give more ablation studies in terms of the architecture, depth of the proposed GFTransformer, as well as the input size of the images. Due to the limited space in the main text, we show more qualitative results to verify the effectiveness of the proposed method in Sec.\ \ref{more experimental results}. Finally, we discuss the ethics statement for our proposed dataset in Sec.\ \ref{ethics statement}.

\section{Implementation Details}
\label{implementation details}
To prepare the grid image features as visual inputs, we use the ResNet-152\cite{he2016deepresnet} pre-trained on the ImageNet\cite{imagenet}, so that these input images are represented as grid features. For region feature extraction in ablation study, we follow \cite{anderson2018vqamethod6} and adopt Faster-RCNN\cite{fasterrcnn} with ResNet-101\cite{he2016deepresnet} backbone pre-trained on the Visual Genome \cite{vg}. The images are represented as a dynamic number (from 10 to 100) of 2048-D region features. For the input language feature extraction, there are two strategies in this paper: the first one is that we just follow the similar settings to \cite{yu2019mcan} based on the 300-D GloVe word embeddings \cite{pennington2014glove} of the tokenized words from the input question, and the second one is that we adopt Bert \cite{devlin2018bert} to encode the input question, which is a deep bidirectional language representation model. The dimension of input question features is set to 512. Both strategies follow a LSTM layer \cite{hochreiter1997lstm} to pre-process the lingual features. For the regularisation part, we apply the BCE loss to learn from the training samples.

In the experiments for benchmark results, we adopt the same hyper-parameters for all transformer networks \cite{martins2020sparse, yu2019mcan, zhou2021trar, aoa, gupta2022swapmix}. As the CNN models \cite{lobry2020rsvqa, yuan2022easy, ben2017mutanmethod10, yang2016stackedsan} do not adopt the similar backbone with ours, we just follow the original network settings in their papers. Note that it is different between RSVQA \cite{lobry2020rsvqa} and HRVQA in terms of the structure and contents of the questions, we just measure their difficulties by using the lengths of the questions for \cite{yuan2022easy}. As for Swap-m \cite{gupta2022swapmix}, it is used as a data augmentation strategy in the comparative experiments. Specifically, it contains two different swapping: context class and context attributes. Due to the different distribution and architecture of the questions between their dataset \cite{hudson2019gqa} and HRVQA, In our settings for benchmark, we only adopt context class swapping for the first 7 question types because the last 3 question types do not contain a specific concept in the question texts, and for the context attribute swapping, only the common attributes ($ size\ and\ location$) are selected for the data augmentation. The training parameters are adopted in all the experiments unless specified otherwise.

\section{Dataset Construction}
\label{dataset construction}
\begin{figure*}[ht]
  \centering
  \includegraphics[width=1\linewidth]{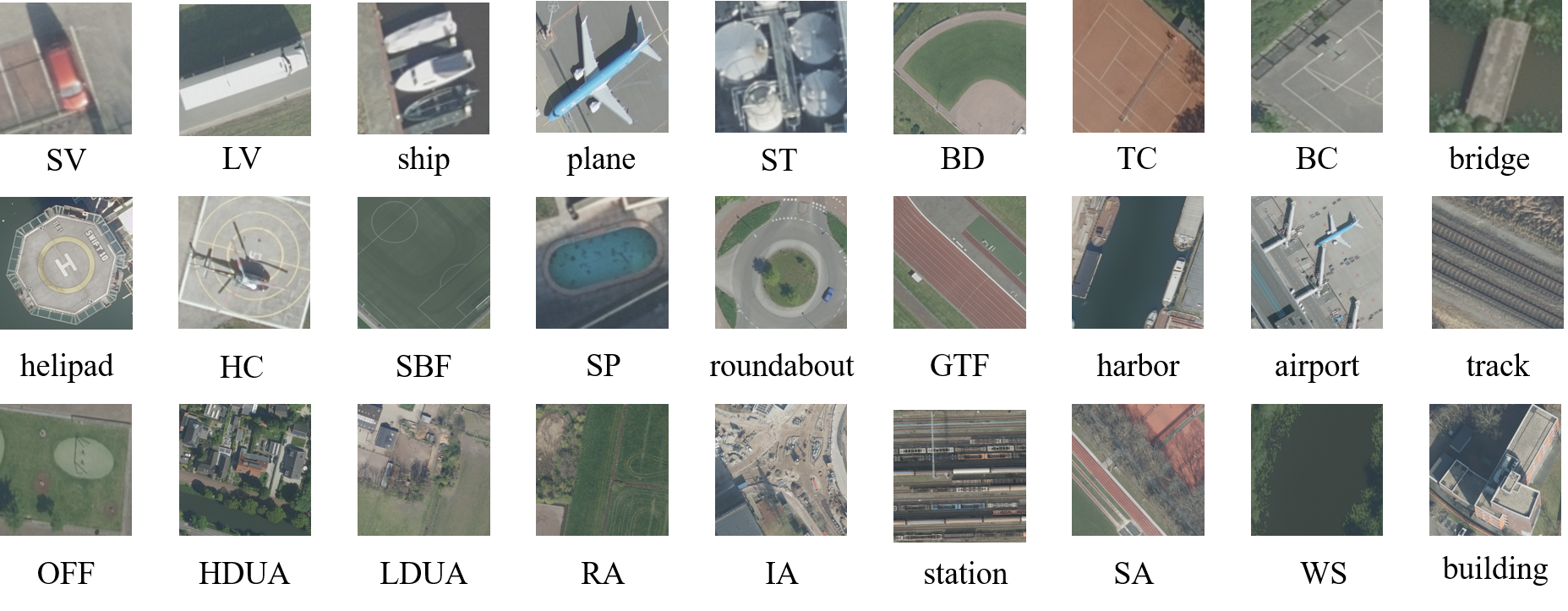}
  \caption{Visual samples of the concepts in HRVQA. The short names for the concepts are defined as: \textit{SV-small vehicle, LV-large vehicle, ST-storage tank, BD-baseball diamond, TC-tennis court, BC-basketball court, HC-helicopter, SBF-soccer ball field, SP-swimming pool, GTF-ground track field, OFF-outdoor fitness field, HDUA-high density urban area, LDUA-low density urban area, RA-rural area, IA-industrial area, SA-sports area, and WS-water section}.}
  \label{fig:samples}
\end{figure*}

\noindent
\textbf{Language Settings.} It is very important to make the input texts clear for a VQA system so that it could generate accurate and meaningful answers. Concepts are the key components in the questions for the language understanding in VQA. Considering the frequency and importance, the proposed concepts are the most related objects and region stuff in earth observation. As it is hard to distinguish different kinds of road networks and provide corresponding VQA triplets, we select \textit{track} to represent this category. Fig. \ref{fig:samples} shows some visual samples of these concepts in HRVQA. To make the concepts more specific and provide richer descriptions for questions, we propose two common attributes (\textit{size and location}) and two special attributes (\textit{shape and color}). The former adapts to every concept such as “building in the top” and “small water section”, while the latter is only suitable for special concepts such as “blue small vehicle” and “square swimming pool”.


\begin{figure}[t!]
  \centering
  \begin{subfigure}[t]{0.39\linewidth}
  \includegraphics[width=1\linewidth]{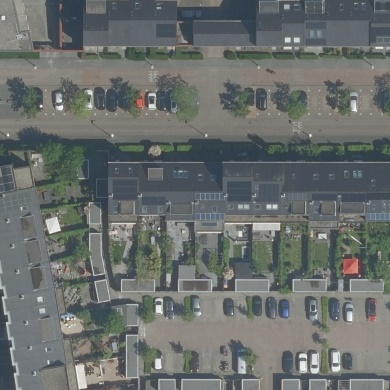}
  \subcaption*{\textbf{$Q$:} What color is the car in the left side?\\\textbf{$A$:} cannot predict}
  \end{subfigure}
  \hfill
  \begin{subfigure}[t]{0.585\linewidth}
  \includegraphics[width=1\linewidth]{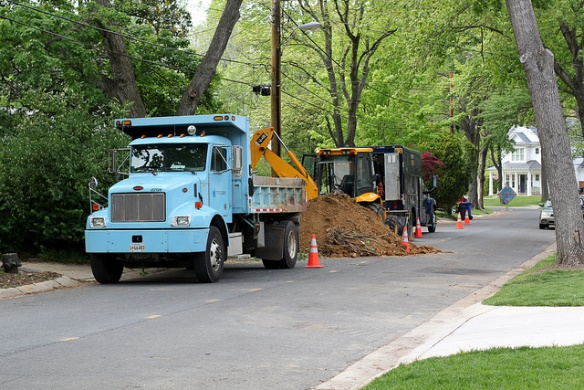}
  \subcaption*{\textbf{$Q$:} What color is the car in the left side?\\\\\textbf{$A$:} blue}
  \end{subfigure}
  \caption{Ambiguity of the same question in different images. \textbf{Left} question is impossible to predict with such ambiguity issue.}
  \vspace{-5mm}
  \label{fig:ambiguity}
\end{figure}

Another important aspect of input texts is the ambiguity issue for aerial image VQA. This problem is caused by the visual appearance in aerial images, which means that high-resolution aerial images contain more various objects than natural images with lower spatial scales. For instance, when the system is asked about the color of a car, the first thing a VQA model should understand is that which one the target is (there are more than ten cars in the \textbf{Left} of Fig. \ref{fig:ambiguity}). In natural images, a general description about the car (such as the car in the left side in the \textbf{Right} of Fig. \ref{fig:ambiguity}) could be enough because the central part of the image only contains no more than three cars and the model could understand it easily. Thus, the complicated visual contents require more detailed texts to describe the intent of the questioner. To make the questions easy-understood without ambiguity in HRVQA, we use a set of more specific descriptions based on the relevant spatial location for the special questions about the semantics of the isolated concepts (e.g., the third car based on left to right rule).


\begin{figure}[t!]
  \centering
  \begin{subfigure}[t]{0.45\linewidth}
  \includegraphics[width=1\linewidth]{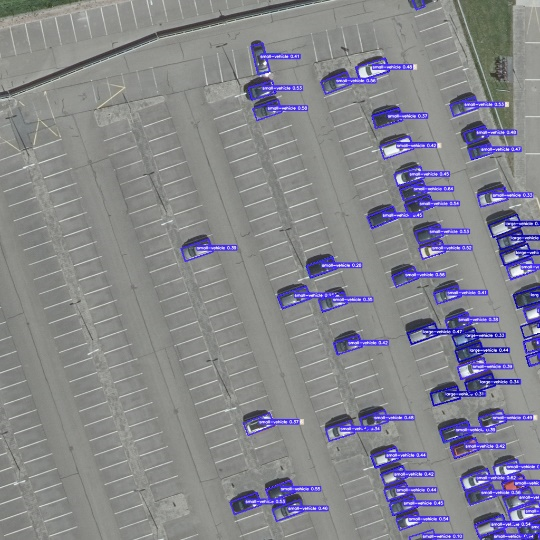}
  \hspace{5mm}
  \includegraphics[width=1\linewidth]{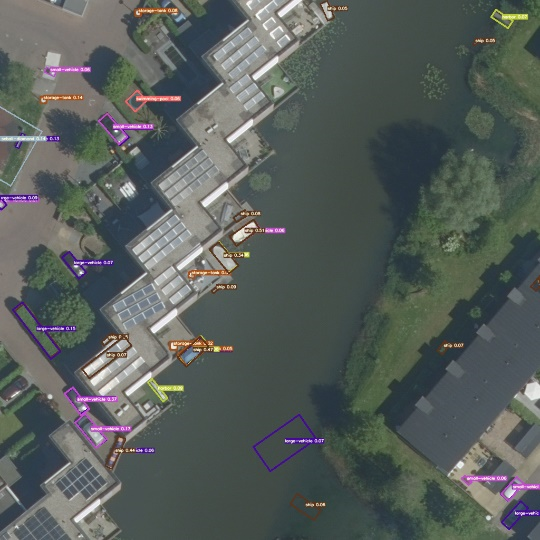}
  \subcaption{Detection results}
  \label{fig:sem-d}
  \end{subfigure}
  \begin{subfigure}[t]{0.45\linewidth}
  \includegraphics[width=1\linewidth]{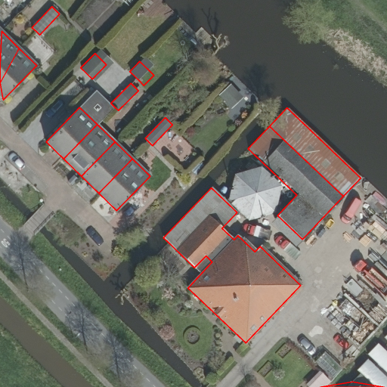}
  \hspace{5mm}
  \includegraphics[width=1\linewidth]{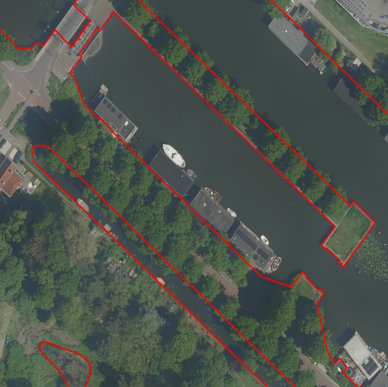}
  \subcaption{Available annotations}
  \label{fig:sem-a}
  \end{subfigure}
  \caption{Extracted semantics of different concepts for labeling.}
  \vspace{-5mm}
  \label{fig:semantics}
\end{figure}

\vspace{2mm}
\noindent
\textbf{Extraction of Semantics.} Based on the pre-trained object detection model, the semantics of some concepts could be extracted from the aerial images to label the image/question pairs. We adopt a powerful object detection network YOLO-v5 \cite{yolov5} in our semi-automatic labeling scheme. The subset of DOTA \cite{dota, dota-v2} (26 aerial images of $7360 \times 4912$ pixels) is used to fine-tune the model, and then this transferred model makes detection inference for the aerial images from HRVQA. Fig. \ref{fig:sem-d} shows the detection results of the fine-tuned model. The Dutch Opendata \cite{pdok} provides the point list annotations for several concepts. To uniformly process semantics from different sources, we transform the polygons to bounding box annotations for objects and keep the pixel statistics based on morphology operations for region stuff. Fig. \ref{fig:sem-a} shows the samples of available annotations from \cite{pdok}.

\begin{table*}
  \caption{Ablation studies for different numbers of layers in GFTransformer with stacked and encoder-decoder architecture, respectively. The short names for the question types are defined as: $Y/N-Yes/No, Locat.-Location, Trans.-Transportation$.}
  \centering
  \setlength\tabcolsep{5pt}
  \begin{tabular}{p{2.5cm}cccccccccccc}
    \toprule
    Layer-Architecture & Number & Y/N & Areas & Size & Locat. & Color & Shape & Sports & Trans. & Scene & OA & AA \\
    \midrule
    layer 2-Stacked   & 66.51 & 93.34 & 97.74 & 93.16 & 69.71 & \textbf{45.90} & 96.57 & 76.90 & 88.61 & 76.65 & 81.25 & 80.51\\
    layer 4-Stacked & 66.82 & 93.35 & 97.68 & 93.43 & 74.04 & 45.21 & 96.64 & \textbf{78.03} & 89.00 & \textbf{78.22} & \textbf{81.90} & \textbf{81.24} \\
    layer 6-Stacked  & \textbf{67.13} & 93.26 & 98.06 & 93.30 & 74.27 & 44.78 & 96.38 & 77.73 & 88.87 & 77.78 & 81.76 & 81.16\\
    layer 8-Stacked  & 66.30 & 93.20 & \textbf{98.22} & 93.20 & 73.79 & 44.63 & 96.51 & 76.63 & 88.68 & 76.78 & 81.40 & 80.79\\
    layer 2-ED   & 66.49 & 93.21 & 97.85 & 93.12 & 70.24 & 44.99 & 96.48 & 76.99 & 88.59 & 76.72 & 81.17 & 80.47\\
    layer 4-ED & 66.53 & \textbf{93.51} & 97.87 & 93.44 & 72.86 & 45.28 & \textbf{96.81} & 77.64 & 88.38 & 77.17 & 81.50 & 80.95 \\
    layer 6-ED (ours) & 66.50 & 93.32 & 97.11 & \textbf{93.72} & \textbf{74.82} & 45.36 & 96.67 & 77.03 & \textbf{88.87} & 77.30 & 81.71 & 81.07\\
    layer 8-ED  & 66.55 & 93.11 & 97.35 & 92.86 & 74.24 & 44.38 & 96.36 & 76.44 & 88.09 & 77.84 & 81.30 & 80.72\\
    \bottomrule
  \end{tabular}
  \label{tab:ablation-layer in vhrqa}
\end{table*}

\begin{table*}[t]
  \caption{Ablation results of type-wise accuracy on HRVQA with smaller input image size. Note that $\dagger$, $\ddagger$, and $\ast$ represent basic CNN models, attention based CNN models, and transformers, respectively.}
  \centering
  \setlength\tabcolsep{5pt}
  \begin{tabular}{p{2.5cm}cccccccccccc}
    \toprule
    Method & Number & Y/N & Areas & Size & Locat. & Color & Shape & Sports & Trans. & Scene & OA & AA \\
    \midrule
    SAN\cite{yang2016stackedsan}$\ddagger$   & 62.38 & 79.80 & 85.04 & 93.41 & 57.92 & 66.91 & 91.00 & 71.22 & 88.62 & 76.92 & 79.36 & 77.32 \\
    MUTAN\cite{ben2017mutanmethod10}$\ddagger$  & 63.87 & 86.88 & 86.05 & 93.53 & 59.46 & \textbf{69.72} & 93.31 & 65.15 & 80.11 & 73.70 & 77.68 & 77.18 \\
    MCAN\cite{yu2019mcan}$\ast$   & 65.75 & 92.58 & 98.21 & 90.88 & 58.38 & 41.56 & 96.37 & 77.82 & 89.52 & 75.71 & 79.96 & 78.68 \\
    RSVQA\cite{lobry2020rsvqa}$\dagger$ & 63.59 & 86.56 & 77.90 & 93.44 & 58.47 & 69.23 & 90.81 & 62.12 & 84.04 & 70.74 & 77.51 & 75.69 \\
    M-CA\cite{martins2020sparse}$\ast$   & 66.65 & 92.80 & 95.11 & 93.70 & 59.27 & 40.97 & 95.59 & 77.58 & 90.08 & 80.03 & 80.79 & 79.18 \\
    TRAR\cite{zhou2021trar}$\ast$   & \textbf{66.96} & \textbf{93.72} & \textbf{98.57} & 87.43 & 58.93 & 46.55 & \textbf{96.63} & 77.23 & 89.66 & 80.25 & 81.15 & 79.59\\
    AOA\cite{aoa}$\ast$ & 66.93 & 93.25 & 97.41 & 93.48 & 60.50 & 45.14 & 96.55 & 78.36 & 89.75 & 79.73 & 81.42 & 80.11 \\
    FETH\cite{yuan2022easy}$\dagger$ & 64.23 & 87.86 & 88.37 & 93.23 & 59.62 & 68.85 & 91.51 & 67.14 & 85.27 & 72.29 & 79.05 & 77.84 \\
    Swap-m\cite{gupta2022swapmix}$\ast$ & 63.45 & 89.88 & 93.42 & 91.27 & 56.81 & 43.55 & 93.55 & 74.32 & 87.27 & 76.93 & 78.56 & 77.04 \\
    GFT ($512 \times 512$)$\ast$   & 65.84 & 92.31 & 97.86 & \textbf{93.77} & \textbf{60.88} & 45.60 & 96.27 & \textbf{78.97} & \textbf{90.42} & \textbf{81.79} & \textbf{81.85} & \textbf{80.37}\\
    \bottomrule
  \end{tabular}
  \label{tab:smaller size results in vhrqa}
\end{table*}

\begin{figure*}[ht]
  \centering
  \begin{subfigure}[t]{0.23\linewidth}
  \includegraphics[width=0.95\linewidth]{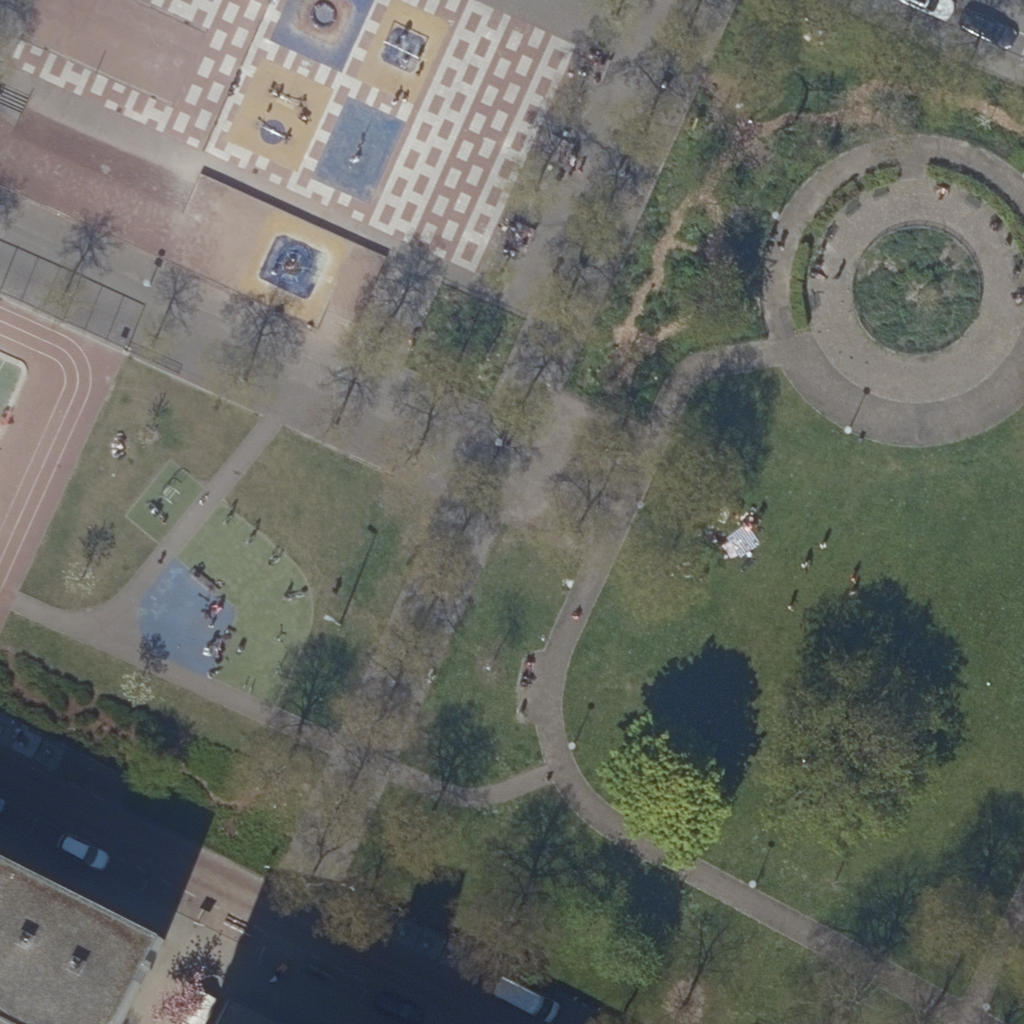}
  \subcaption*{\textbf{$Q$:} What kind of size is the first outdoor fitness field based on the left to right rule?\\ \textbf{$A$:} large scale}
  \end{subfigure}
  \begin{subfigure}[t]{0.23\linewidth}
  \includegraphics[width=0.95\linewidth]{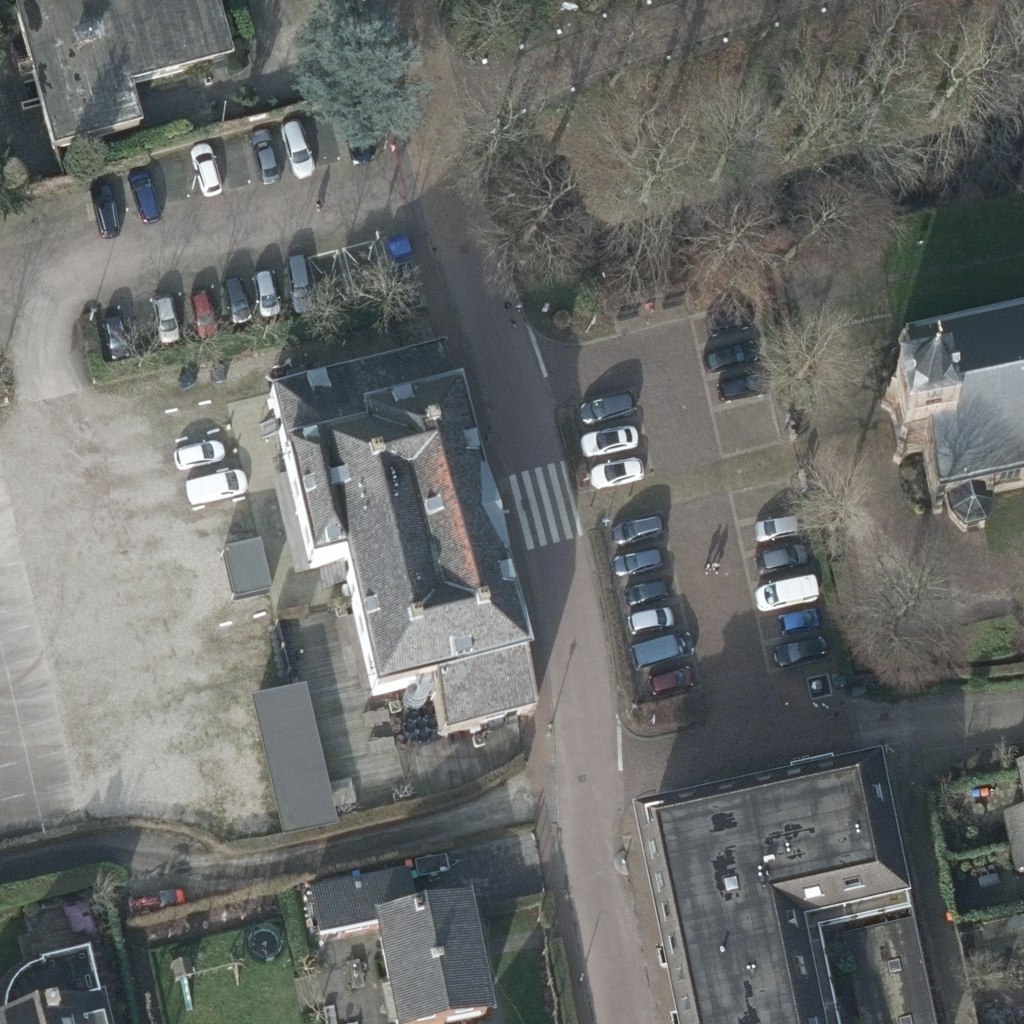}
  \subcaption*{\textbf{$Q$:} What is the color of the third small vehicle based on the left to right rule in this image?\\ \textbf{$A$:} blue}
  \end{subfigure}
  \begin{subfigure}[t]{0.23\linewidth}
  \includegraphics[width=0.95\linewidth]{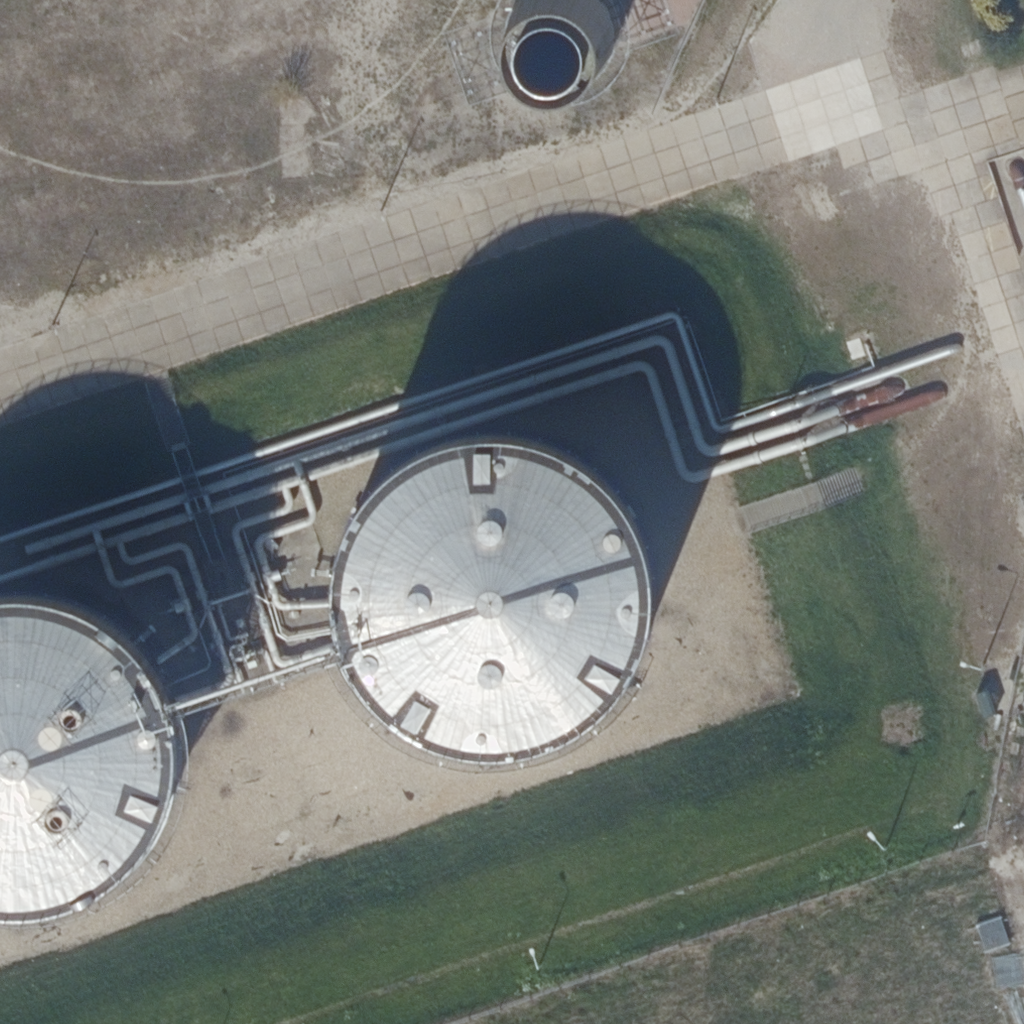}
  \subcaption*{\textbf{$Q$:} Where is the second storage-tank based on the left to right rule in this image?\\ \textbf{$A$:} center}
  \end{subfigure}
  \begin{subfigure}[t]{0.23\linewidth}
  \includegraphics[width=0.95\linewidth]{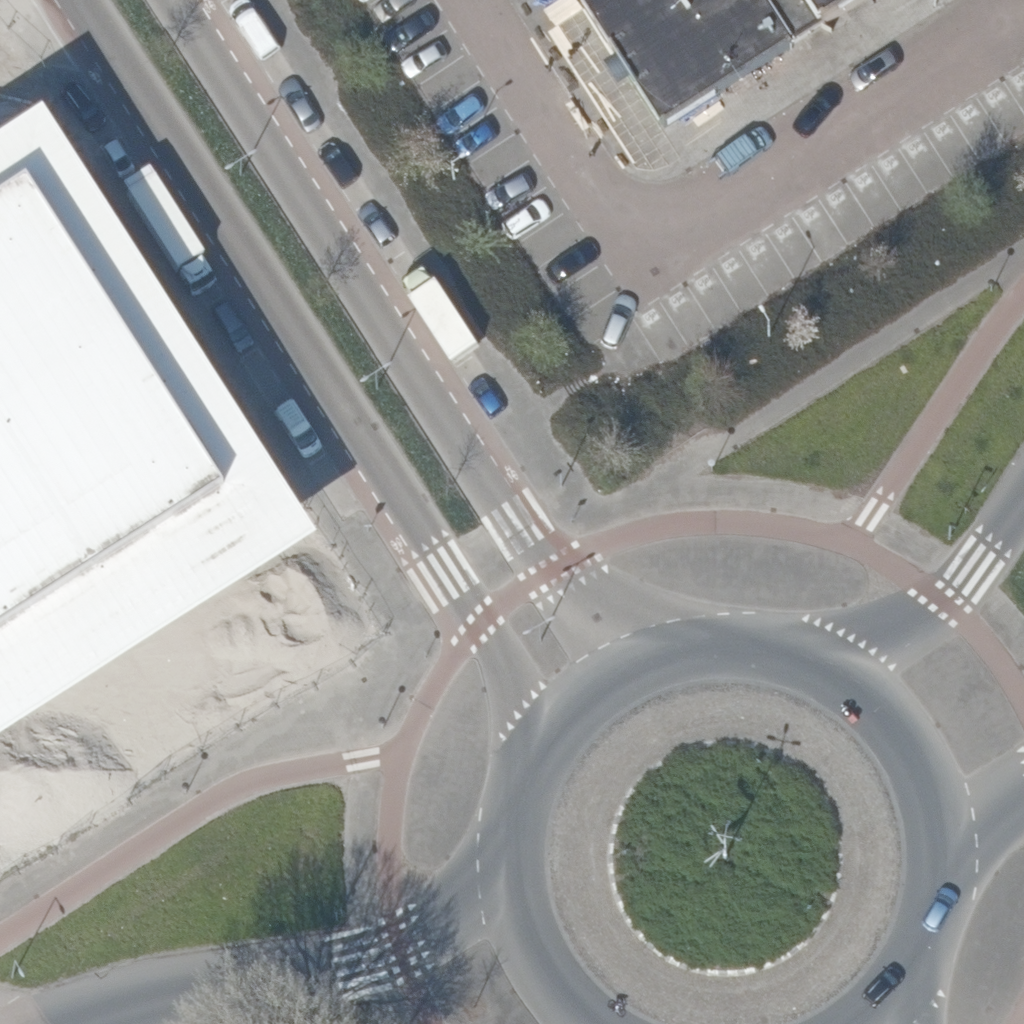}
  \subcaption*{\textbf{$Q$:} What kind of shape is the roundabout based on the left to right rule in this image?\\ \textbf{$A$:} round}
  \end{subfigure}
  \begin{subfigure}[t]{0.23\linewidth}
  \includegraphics[width=0.95\linewidth]{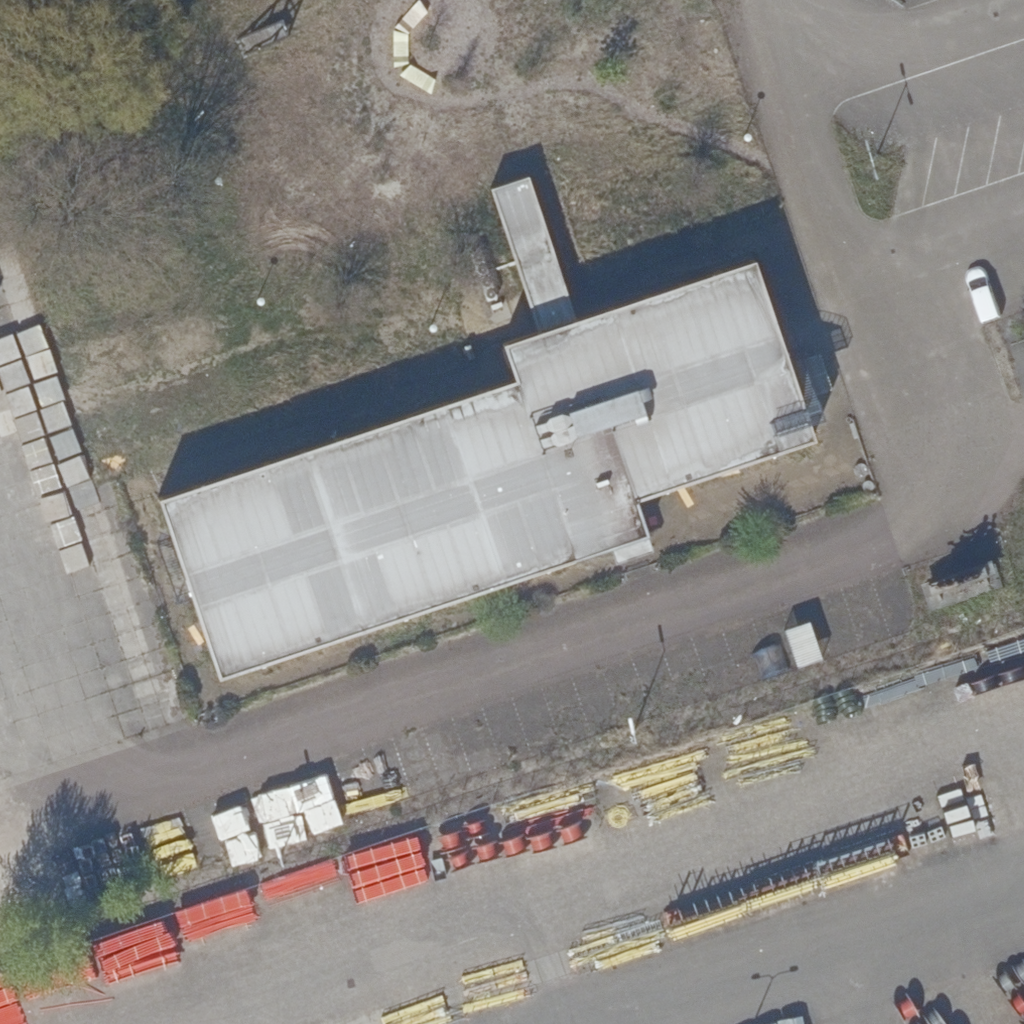}
  \subcaption*{\textbf{$Q$:} Is there any water sections in this image?\\ \textbf{$A$:} no}
  \end{subfigure}
  \begin{subfigure}[t]{0.23\linewidth}
  \includegraphics[width=0.95\linewidth]{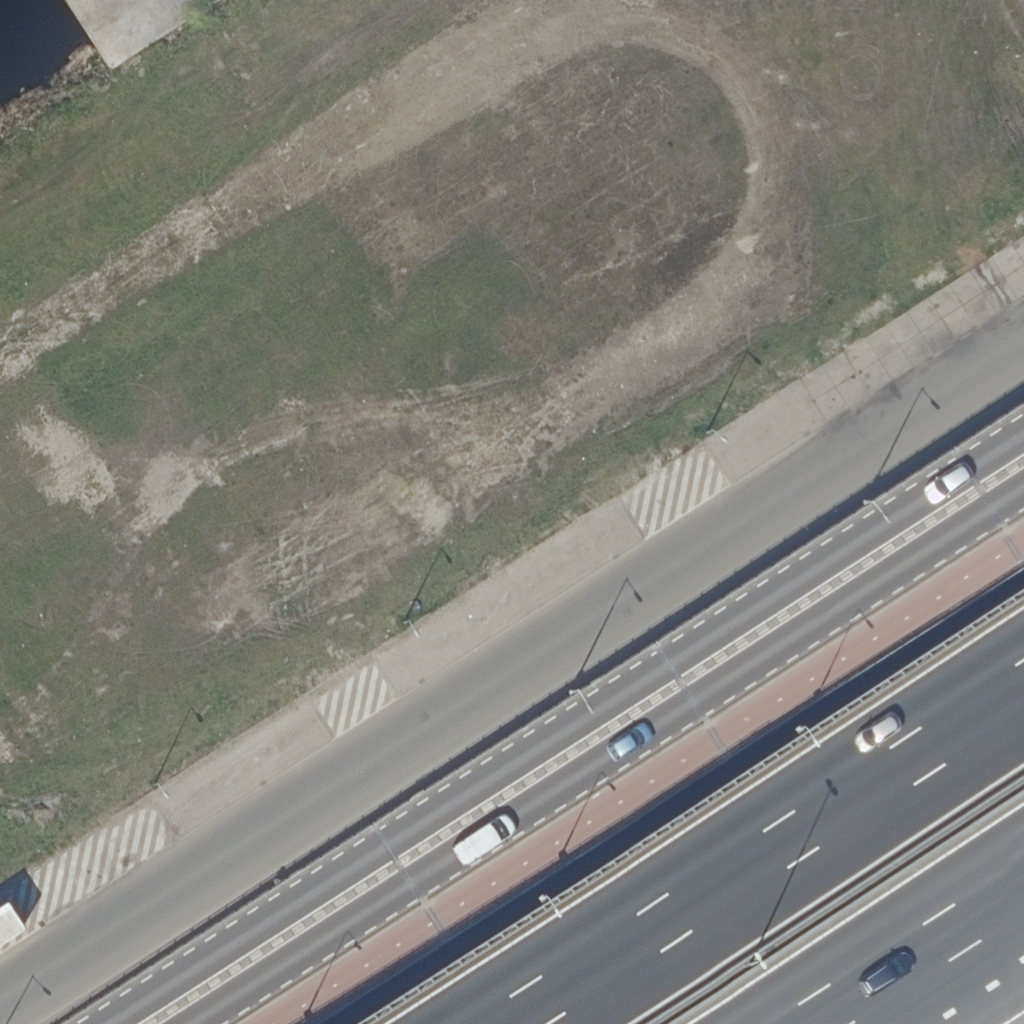}
  \subcaption*{\textbf{$Q$:} How many small vehicles can you see in this image?\\ \textbf{$A$:} 5}
  \end{subfigure}
  \begin{subfigure}[t]{0.23\linewidth}
  \includegraphics[width=0.95\linewidth]{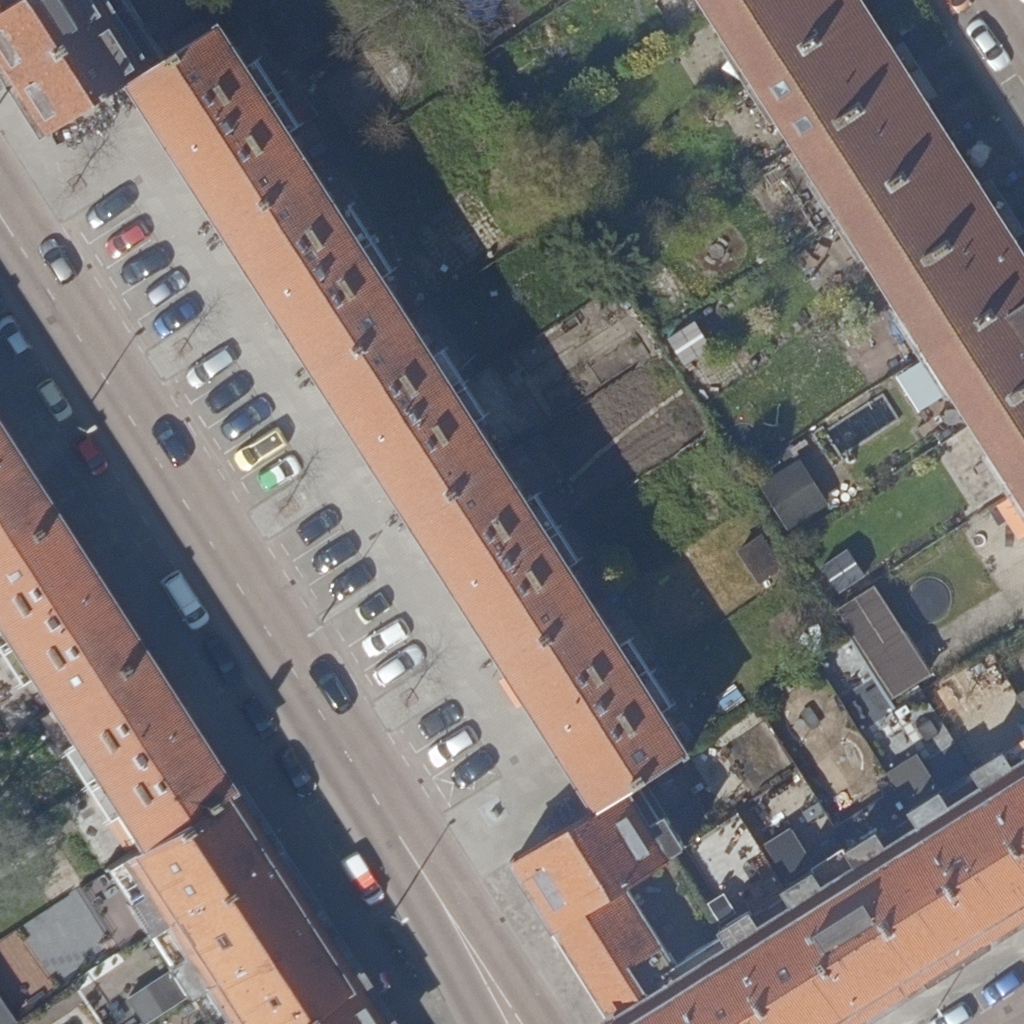}
  \subcaption*{\textbf{$Q$:} What is the main scene of this image?\\ \textbf{$A$:} high density urban area\\}
  \end{subfigure}
  \begin{subfigure}[t]{0.23\linewidth}
  \includegraphics[width=0.95\linewidth]{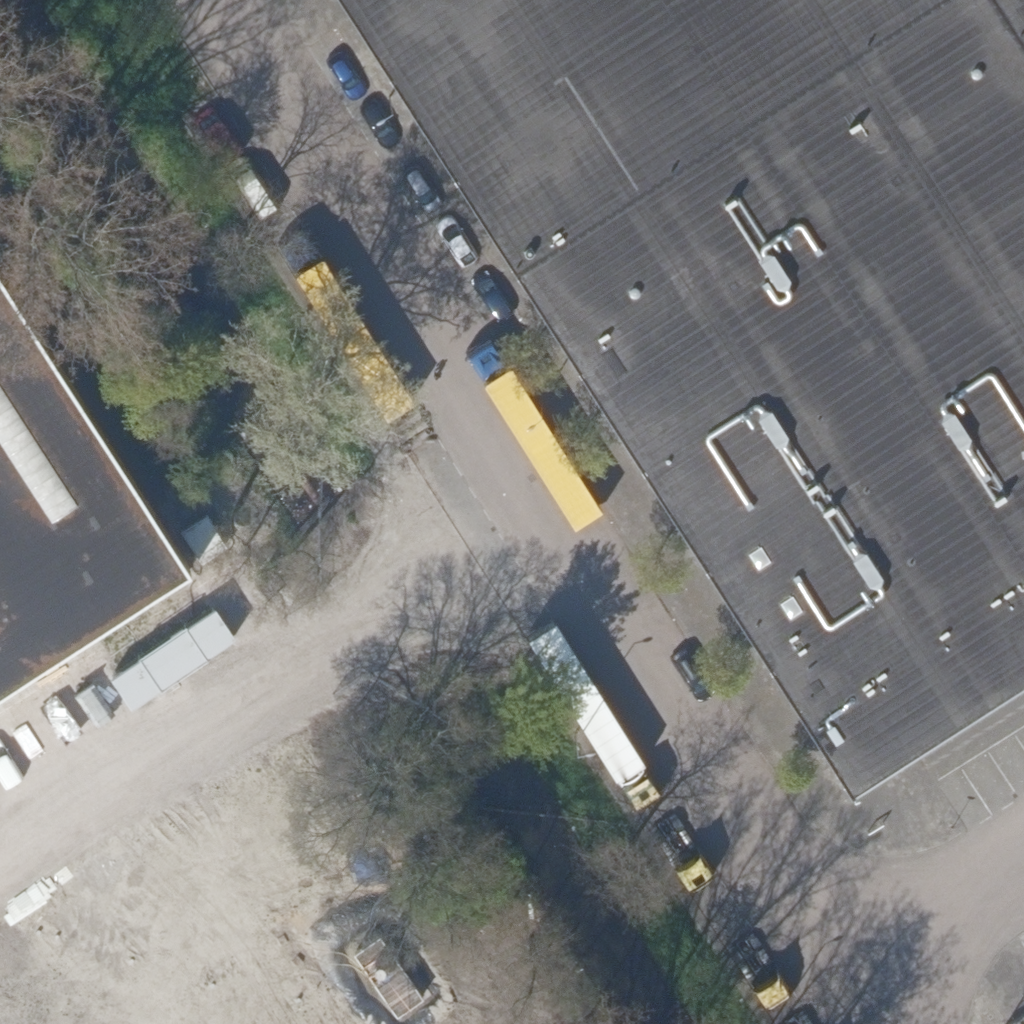}
  \subcaption*{\textbf{$Q$:} Does a building exist in this image?\\ \textbf{$A$:} yes}
  \end{subfigure}
  
  \begin{subfigure}[t]{0.23\linewidth}
  \includegraphics[width=0.95\linewidth]{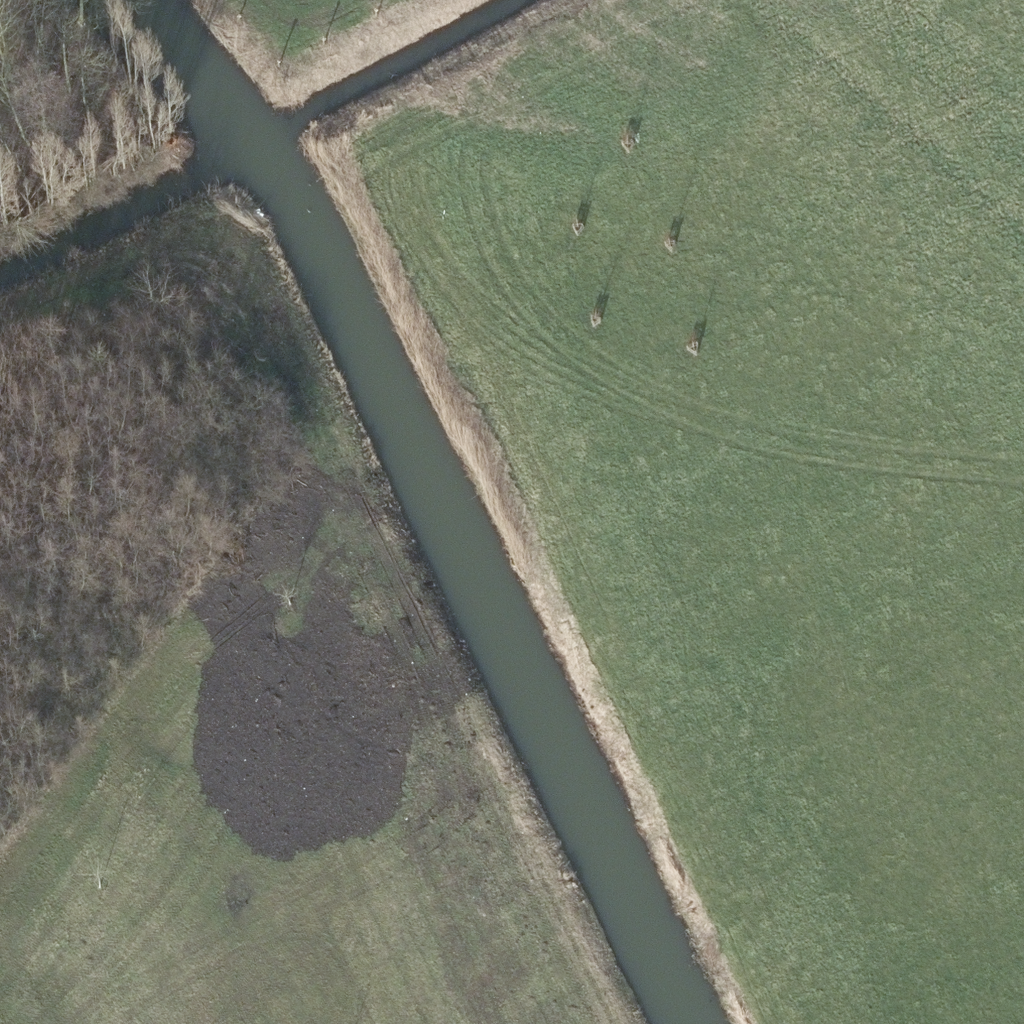}
  \subcaption*{\textbf{$Q$:} Can you see a green ship in this image?\\\\ \textbf{$A$:} no}
  \end{subfigure}
  \begin{subfigure}[t]{0.23\linewidth}
  \includegraphics[width=0.95\linewidth]{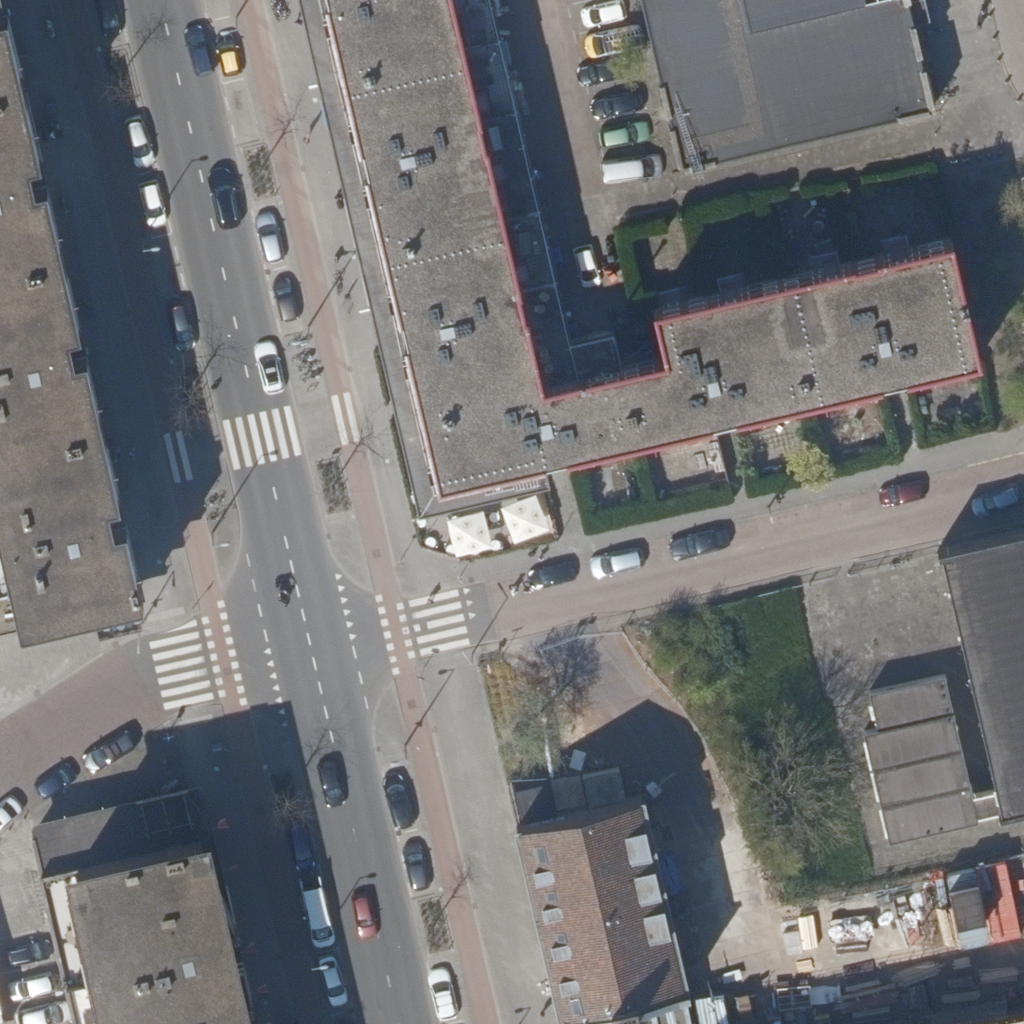}
  \subcaption*{\textbf{$Q$:} What color is the fourth small vehicle based on the left to right rule in this image?\\ \textbf{$A$:} grey}
  \end{subfigure}
  \begin{subfigure}[t]{0.23\linewidth}
  \includegraphics[width=0.95\linewidth]{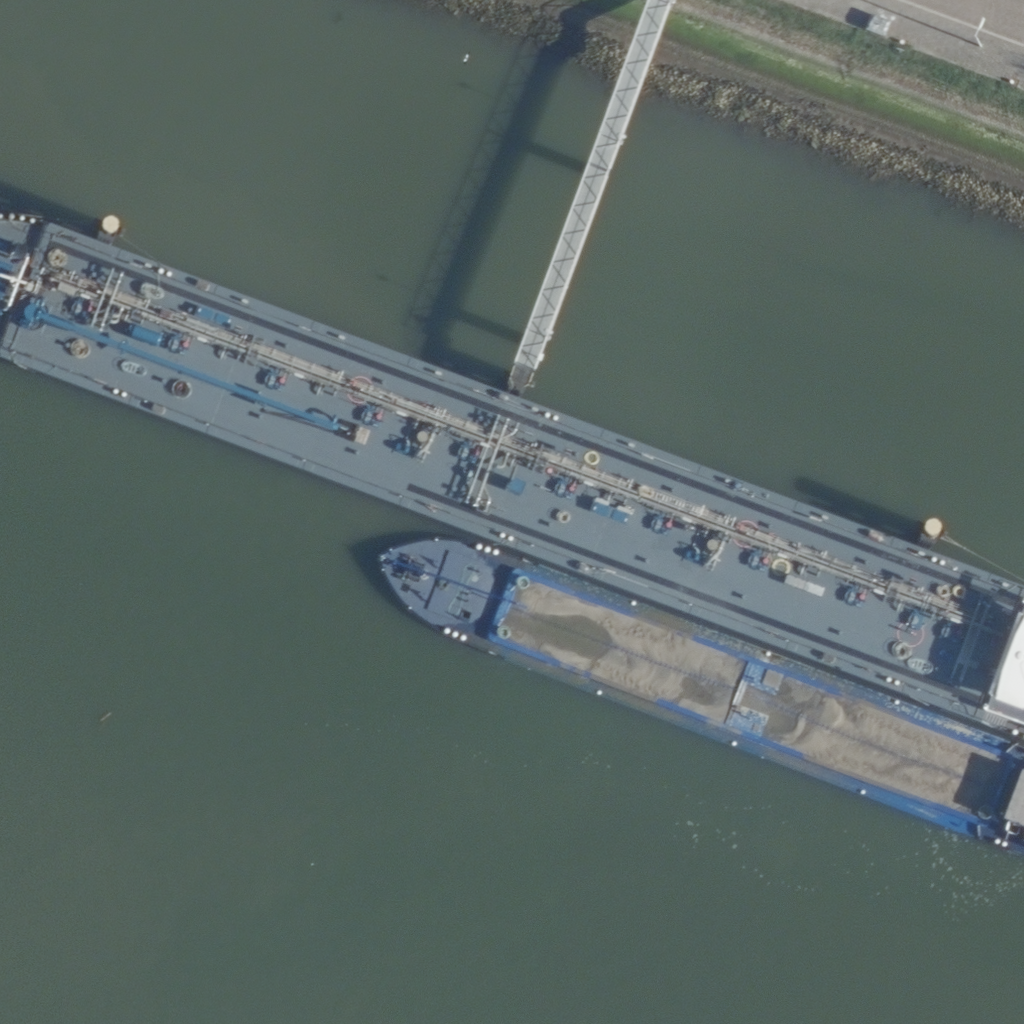}
  \subcaption*{\textbf{$Q$:} What is the main transportation of this image?\\\\ \textbf{$A$:} by ship}
  \end{subfigure}
  \begin{subfigure}[t]{0.23\linewidth}
  \includegraphics[width=0.95\linewidth]{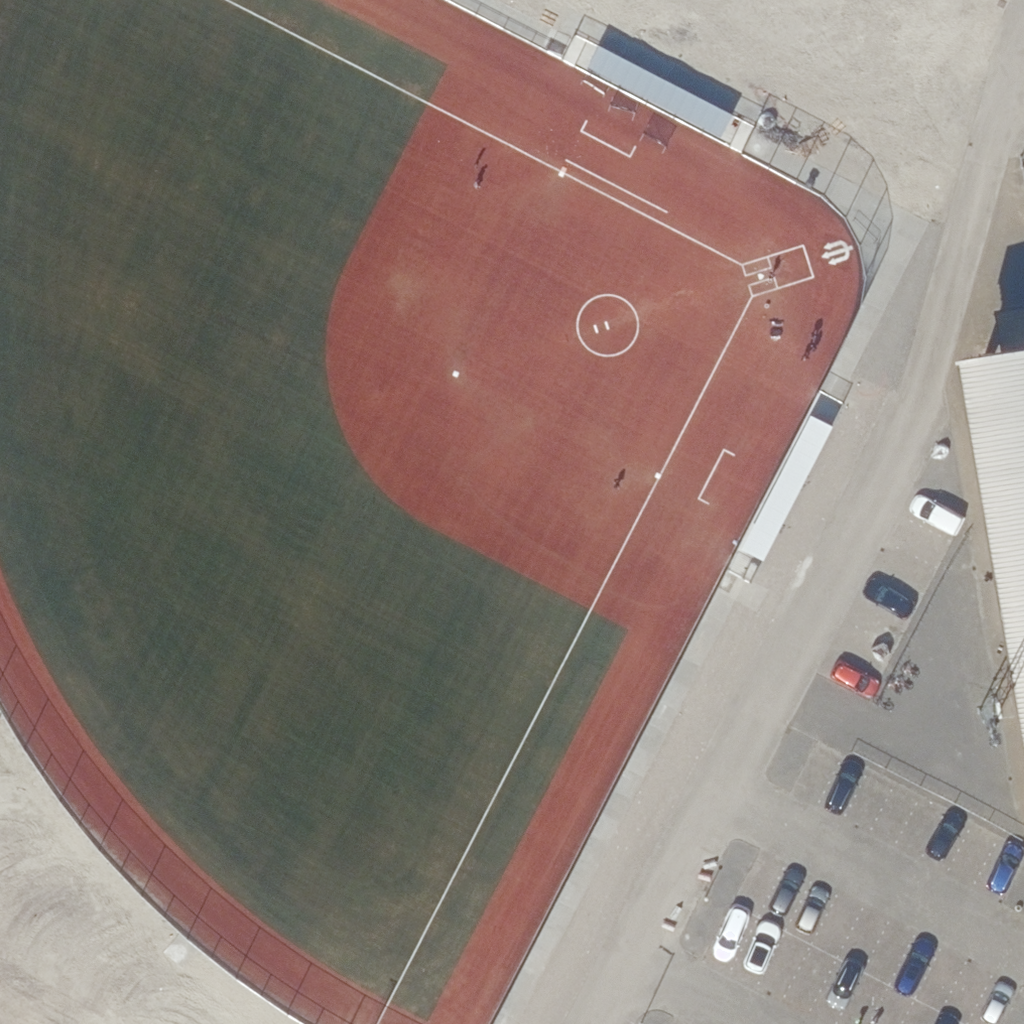}
  \subcaption*{\textbf{$Q$:} What kind of scene can you see in this image?\\\\ \textbf{$A$:} sports area}
  \end{subfigure}
  
  \begin{subfigure}[t]{0.23\linewidth}
  \includegraphics[width=0.95\linewidth]{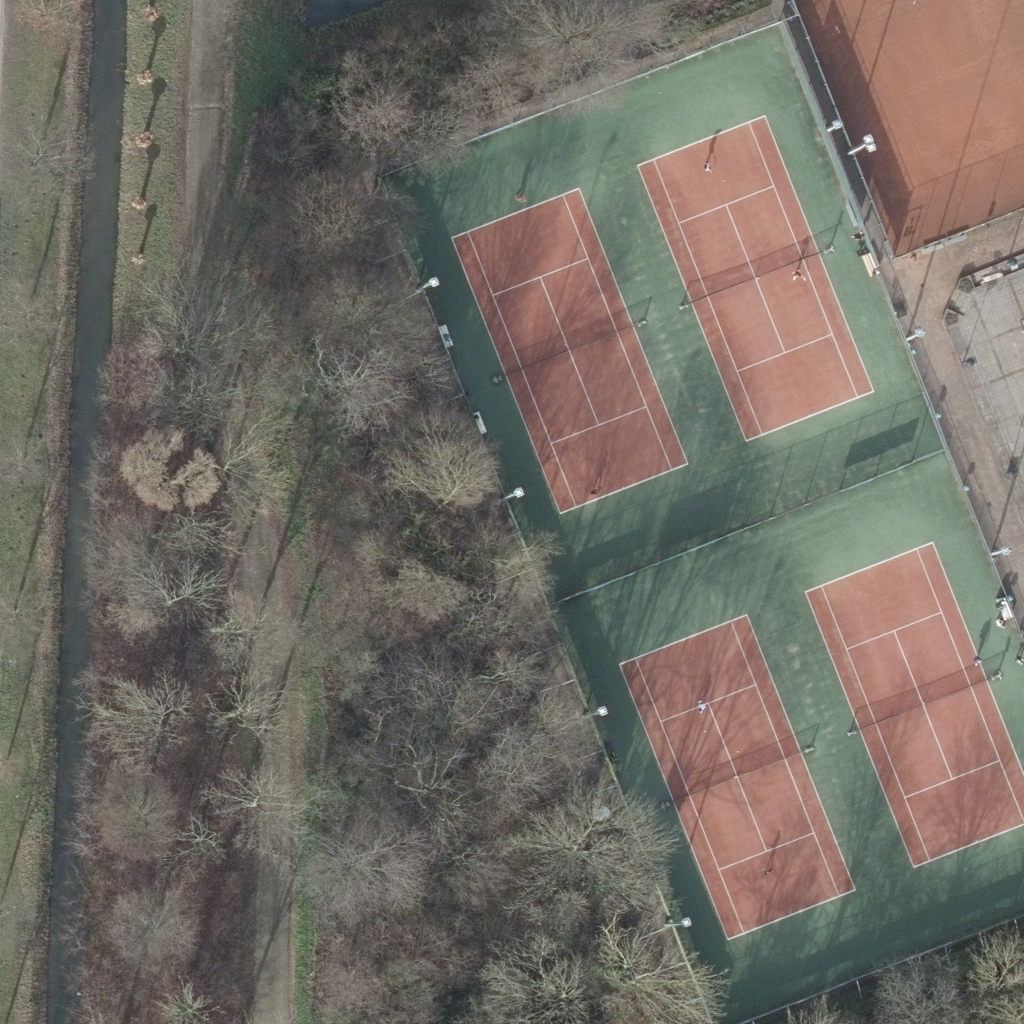}
  \subcaption*{\textbf{$Q$:} What sport can people do in this image?\\ \textbf{$A$:} playing tennis}
  \end{subfigure}
  \begin{subfigure}[t]{0.23\linewidth}
  \includegraphics[width=0.95\linewidth]{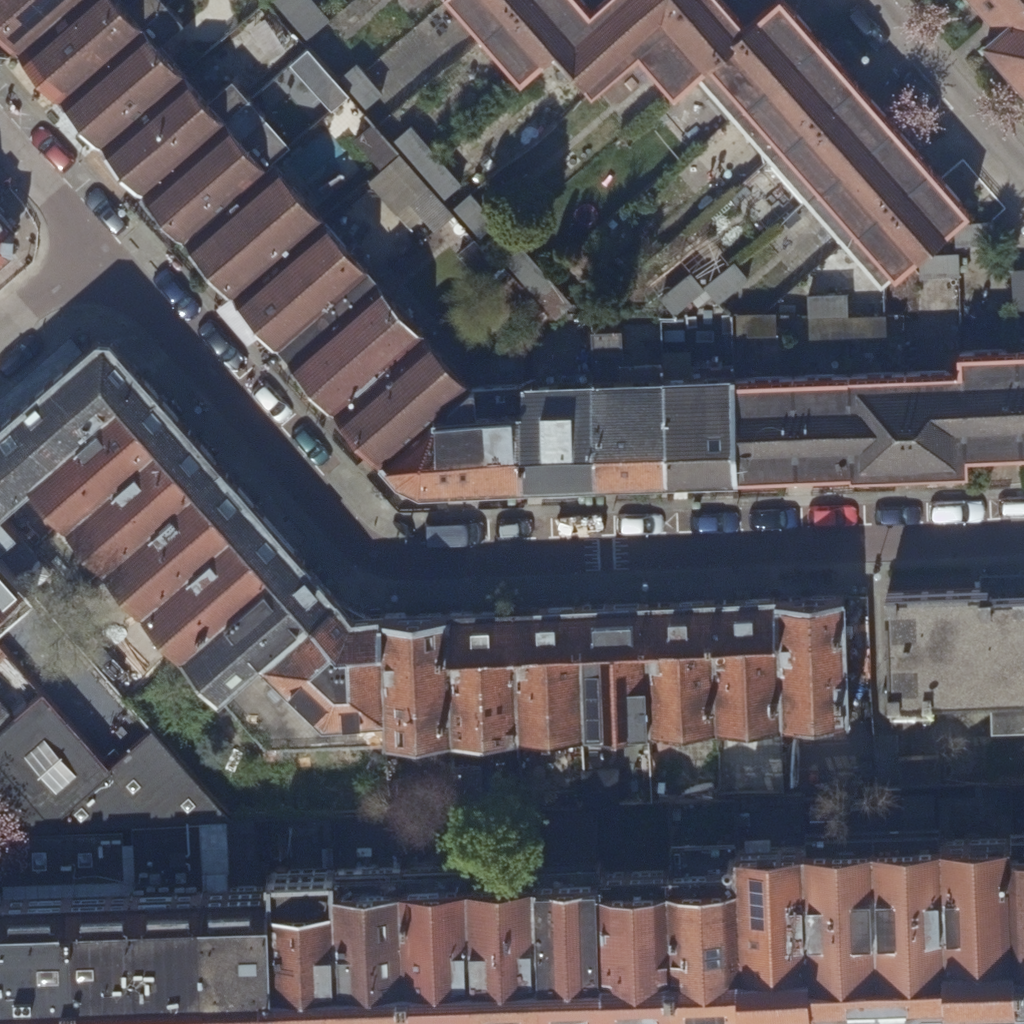}
  \subcaption*{\textbf{$Q$:} How many small vehicles are visible in this image?\\ \textbf{$A$:} 15}
  \end{subfigure}
  \begin{subfigure}[t]{0.23\linewidth}
  \includegraphics[width=0.95\linewidth]{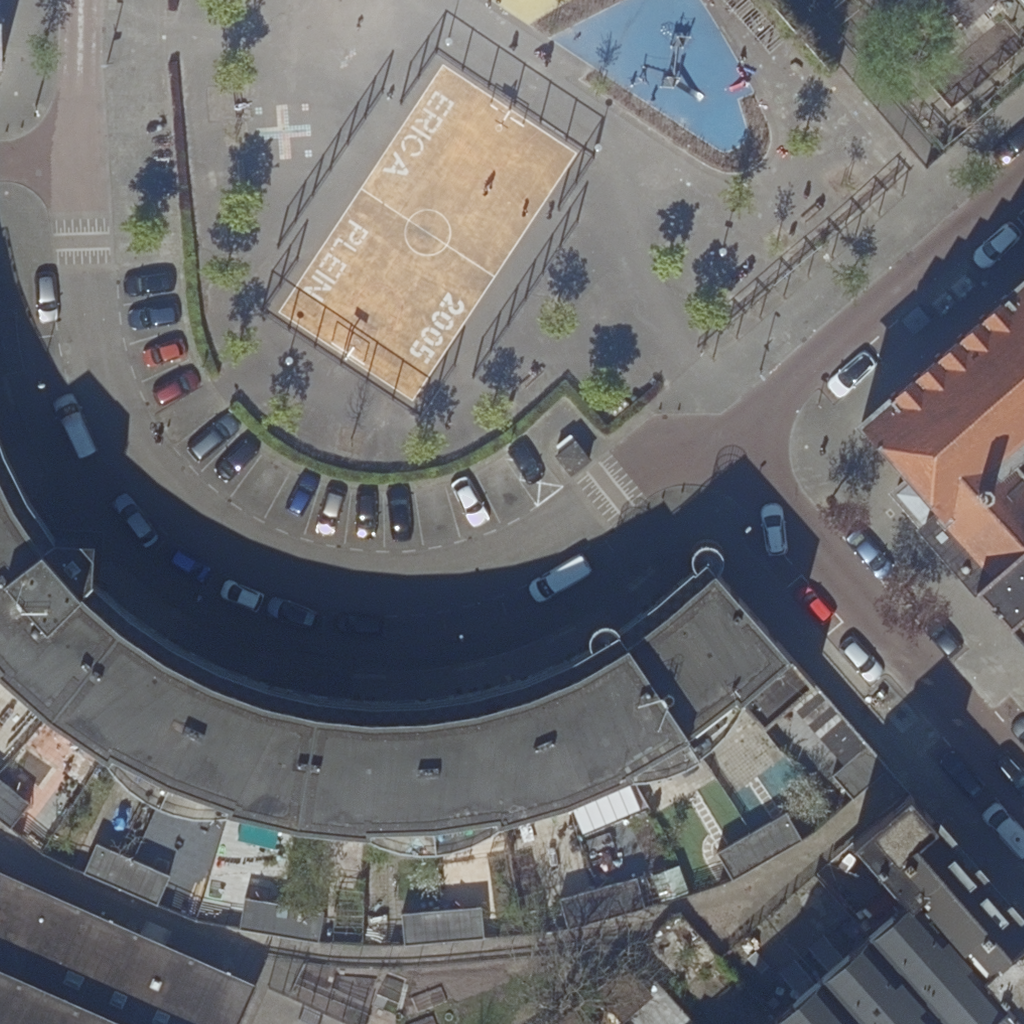}
  \subcaption*{\textbf{$Q$:} Which part of this image is the outdoor fitness field located in?\\ \textbf{$A$:} top right}
  \end{subfigure}
  \begin{subfigure}[t]{0.23\linewidth}
  \includegraphics[width=0.95\linewidth]{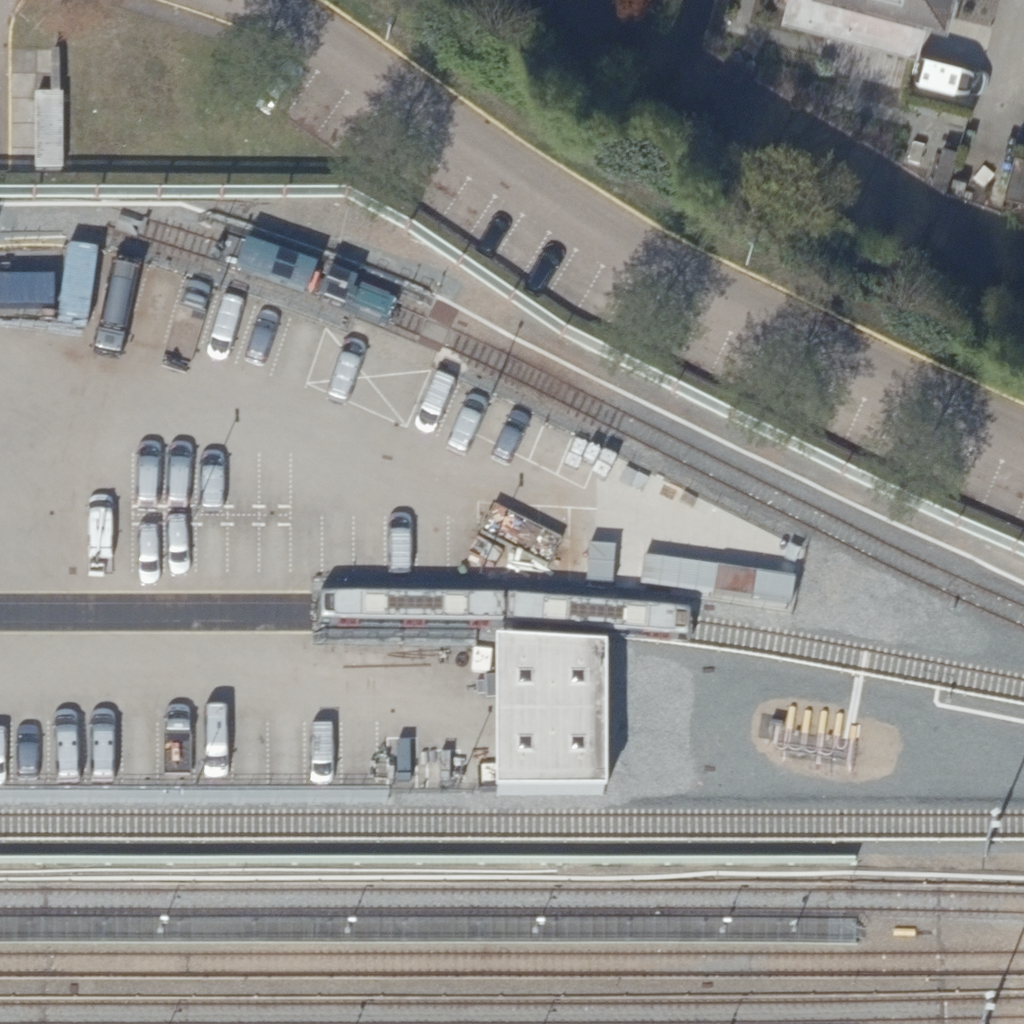}
  \subcaption*{\textbf{$Q$:} What is the area covered by tracks in this image?\\ \textbf{$A$:} more than 1000$m^2$}
  \end{subfigure}
  \caption{Random examples of image/question/answer triplets in HRVQA.}
  \label{fig:additional examples}
\end{figure*}

\vspace{2mm}
\noindent
\textbf{Triplet Construction.} For each image in HRVQA, we randomly generate 200 questions based on the language settings mentioned above. Then based on the concepts, question type, attributes, we search for the semantics in the extracted results to generate the corresponding answers. Specifically, according to the selected question type, common questions and special questions (no specific concepts included in the question texts, e.g., \textit{Scene, Transportation, and Sports}) are separated into different groups. Then for common questions, concepts could be classified into isolated objects and region stuff considering the optional attributes. Isolated objects could be asked with both common and special attributes (\textit{color and shape}) while region stuff is only related to common attributes (\textit{size and location}). After determining the content of the question, a template will be selected from the question reference set. The extracted semantics (spatial and category information) could help generate the corresponding answer to the image/question pair and a flag indicating the relevance. To provide insight into our proposed dataset, we show additional examples of the automatically generated triplets in Fig. \ref{fig:additional examples}.

\begin{figure*}[ht]
  \centering
  \begin{subfigure}[t]{0.48\linewidth}
  \includegraphics[width=1\linewidth]{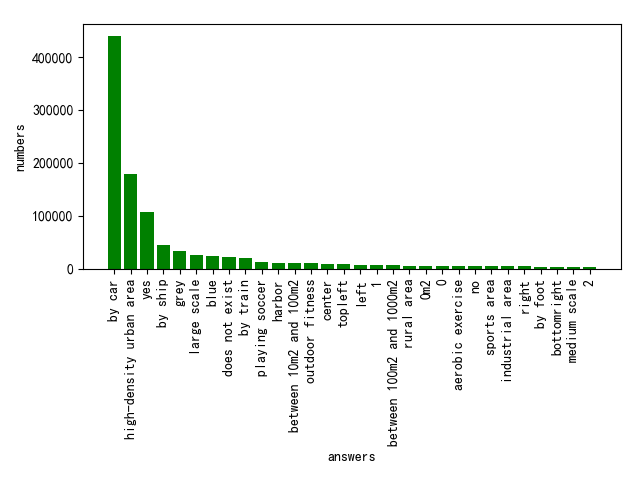}
  \subcaption{Before filtering}
  \label{fig:dis-a1}
  \end{subfigure}
  \hspace{1mm}
  \begin{subfigure}[t]{0.48\linewidth}
  \includegraphics[width=1\linewidth]{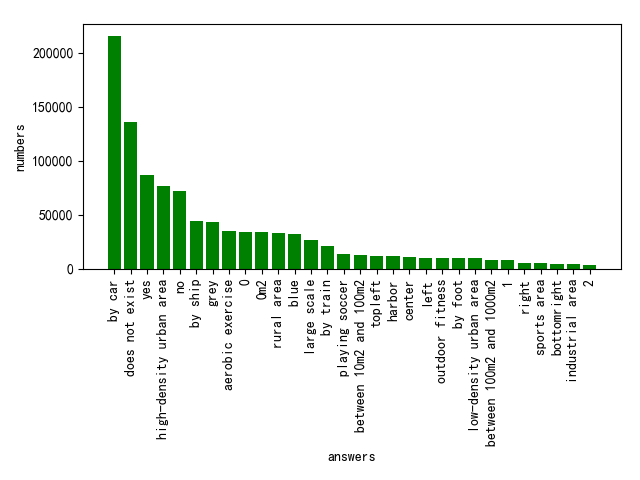}
  \subcaption{After filtering}
  \label{fig:dis-a2}
  \end{subfigure}
  \caption{Answer distribution before and after the filter.}
  \vspace{-3mm}
  \label{fig:answer-dis}
\end{figure*}

\vspace{2mm}
\noindent
\textbf{Triplet Filtering.} The aim of this process is to keep the meaningful triplets from the automatically generated samples and alleviate the influence of the short-cuts. In order to intuitively understand the meaning of the filtering, Fig. \ref{fig:dis-a1} and \ref{fig:dis-a2} compare the top 30 answer distribution before and after filtering, respectively. It shows that the filter does remove irrelevant answers but still leaves a long-tail issue for urban scenes. Besides, class imbalance and label noise problems are unavoidable in our semi-automatically constructed dataset. 
This issue requires more explorations in the future research.

\begin{figure}[ht]
  \centering
  \begin{subfigure}[t]{0.48\linewidth}
  \includegraphics[width=1\linewidth]{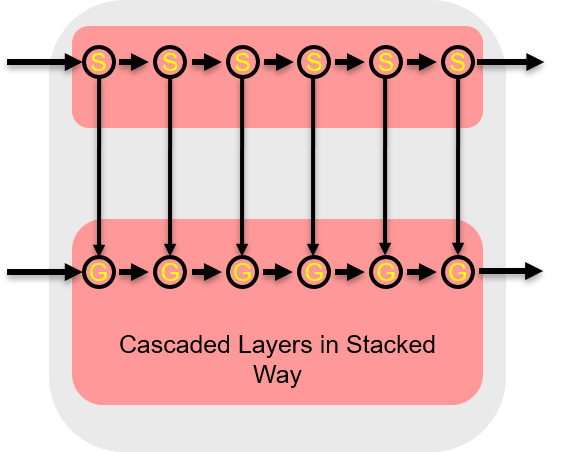}
  \subcaption{Stacked}
  \end{subfigure}
  \hfill
  \begin{subfigure}[t]{0.48\linewidth}
  \includegraphics[width=1\linewidth]{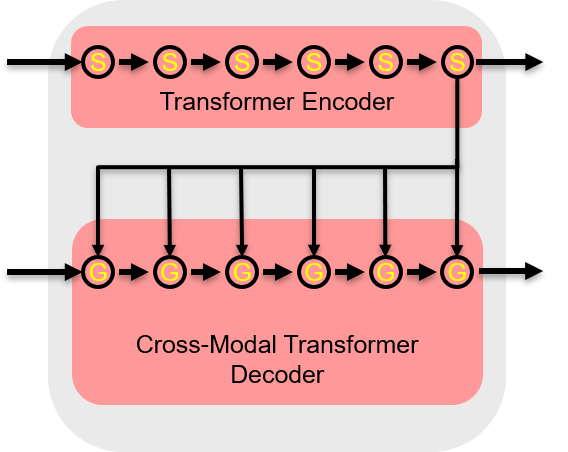}
  \subcaption{Encoder-decoder}
  \end{subfigure}
  \caption{Cascaded layers in stacked and encoder-decoder architecture, respectively.}
  \vspace{-1mm}
  \label{fig:stacked and ed}
\end{figure}

\section{More Ablation Studies}
\label{more ablation studies}
In order to demonstrate the impact of our proposed GFTransformer in terms of architecture, depth of network, and input size, we conduct additional ablation studies and report the corresponding results in this section.

\vspace{2mm}
\noindent
\textbf{Different architecture.} In our proposed GFTransformer, we feed the network by using an encoder-decoder architecture, which means the lingual features are completely refined based on the self-attention units before guiding the other modality in guided attention units. Another popular architecture for deep learning is stacking the layers in depth, which means the lingual features and visual features are refined together in each individual layer with the increasing depth. Fig. \ref{fig:stacked and ed} shows these two different architecture for cross-modal learning in our VQA task. Table \ref{tab:ablation-layer in vhrqa} report the results on HRVQA, and it shows that stacked architecture performed a little bit better than encoder-decoder. This is because the learned self-attentions from lingual features are not so stable without mutual guidance compared to that from stacked layers, especially the questions about general information from the input images (e.g., \textit{Sports, Transportation and Scene}). Furthermore, the core words in these question are always the functions of the text (e.g., transportation, scene), then the last self-attentions from lingual features are not so needed compared to that from the step-by-step guidance in the refining of visual features. Due to the larger proportions of these questions in HRVQA, they lead a little bit better results in OA and AA. As the key difficulty of this dataset lies in the questions about the specific attributes of targets and the encoder-decoder architecture performs better on these questions, we insist the encoder-decoder network in our proposed GFTransformer for HRVQA.

\vspace{2mm}
\noindent
\textbf{Depth of Network.} In the last 4 rows of Table \ref{tab:ablation-layer in vhrqa}, we report the results based on different number of layers in depth for encoder-decoder architecture. Specifically, with increasing number \textit{layer}, the performance first rise and then fall when the \textit{layer=6}. Interestingly, we find the similar phenomenon in other works \cite{aoa}: the increase of the network depth does not necessarily bring about the improvement of the accuracy, but will probably impair the performance of the models. In addition to the results in encoder-decoder architecture, the similar phenomenon happens in stacked architecture, as shown in the first 4 rows of Table \ref{tab:ablation-layer in vhrqa}. Thus, we choose 6 as our layer depth in GFTransformer.

\vspace{2mm}
\noindent
\textbf{Size of Images.} As the size of images for previous RSVQA related methods \cite{lobry2020rsvqa,yuan2022easy} are set 512, we conduct additional experiments with the same size for other models including ours, to make a fair comparison. We simply resize the original patches since the labeling for question/answer is extremely expensive. In this part, we run the comparative ablations on the resized images in $512 \times 512$, as shown in Table \ref{tab:smaller size results in vhrqa}. It shows that with smaller input size, the models handle overall information (information that needs to analyze the whole images from an overall perspective, e.g., scene) better while achieving worse results on the specific locations of the concepts.

\section{More Experimental Results}
\label{more experimental results}
In this section, we provide more qualitative results to verify the effectiveness of the proposed method, including some failure cases. 

\begin{figure*}[ht]
  \centering
  \begin{subfigure}[t]{0.23\linewidth}
  \includegraphics[width=0.95\linewidth]{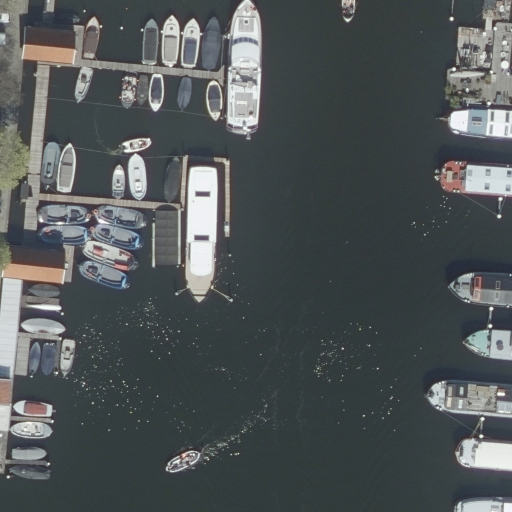}
  \subcaption*{\textbf{$Q$:} What kind of scene can you see in this image?\\ \textbf{$A_{GT}$:} harbor\\ \textbf{$A_P$:} harbor}
  \end{subfigure}
  \begin{subfigure}[t]{0.23\linewidth}
  \includegraphics[width=0.95\linewidth]{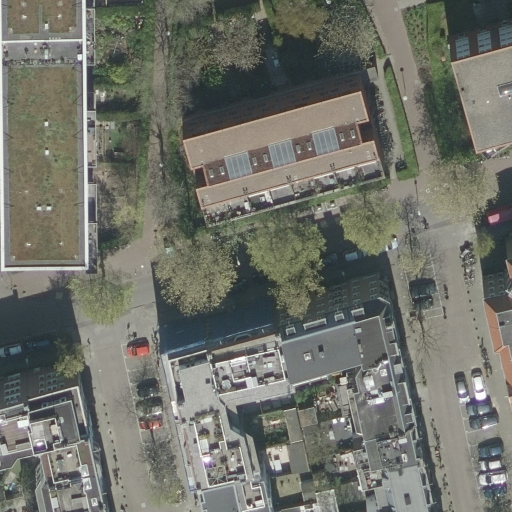}
  \subcaption*{\textbf{$Q$:} Is there a green small vehicle in this image?\\ \textbf{$A_{GT}$:} yes\\ \textbf{$A_P$:} yes}
  \end{subfigure}
  \begin{subfigure}[t]{0.23\linewidth}
  \includegraphics[width=0.95\linewidth]{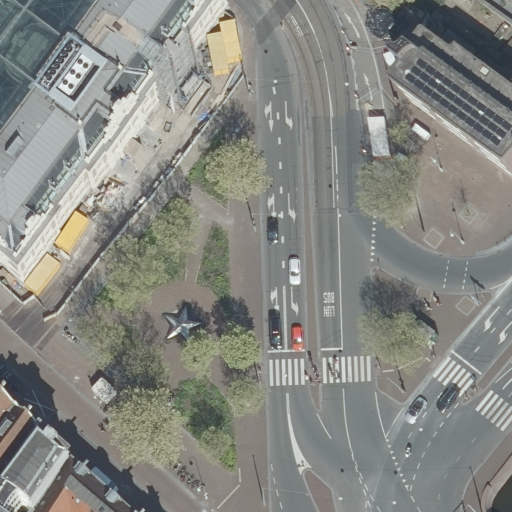}
  \subcaption*{\textbf{$Q$:} What is the main transportation of this image?\\ \textbf{$A_{GT}$:} by car\\ \textbf{$A_P$:} by car}
  \end{subfigure}
  \begin{subfigure}[t]{0.23\linewidth}
  \includegraphics[width=0.95\linewidth]{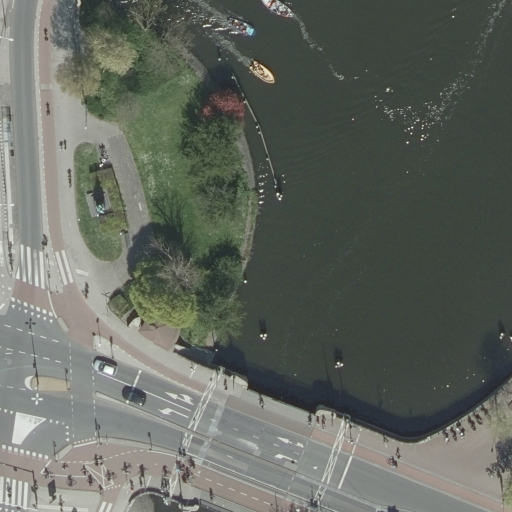}
  \subcaption*{\textbf{$Q$:} How many small vehicles are there in this image?\\ \textbf{$A_{GT}$:} 2\\ \textbf{$A_P$:} 2}
  \end{subfigure}
  
  \begin{subfigure}[t]{0.23\linewidth}
  \includegraphics[width=0.95\linewidth]{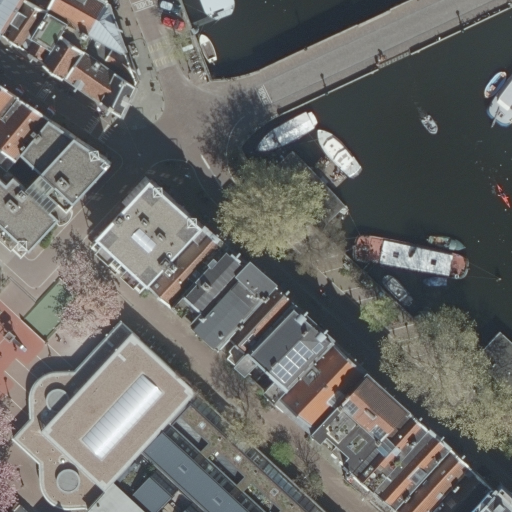}
  \subcaption*{\textbf{$Q$:} What size is the first bridge based on the left to right rule in this image?\\ \textbf{$A_{GT}$:} large scale\\ \textbf{$A_P$:} large scale}
  \end{subfigure}
  \begin{subfigure}[t]{0.23\linewidth}
  \includegraphics[width=0.95\linewidth]{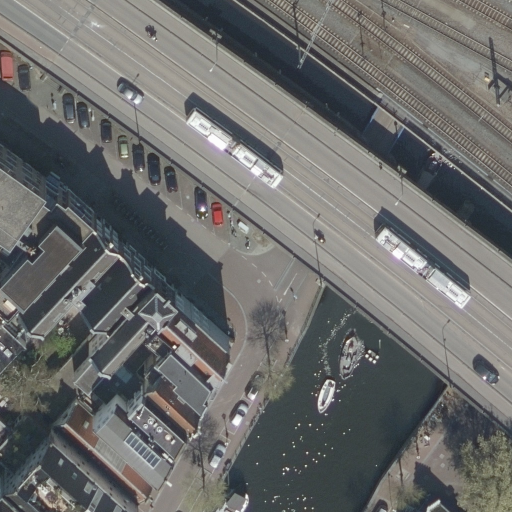}
  \subcaption*{\textbf{$Q$:} Which part of this image is the second ship based on the left to right rule located?\\ \textbf{$A_{GT}$:} bottom right\\ \textbf{$A_P$:} bottom right}
  \end{subfigure}
  \begin{subfigure}[t]{0.23\linewidth}
  \includegraphics[width=0.95\linewidth]{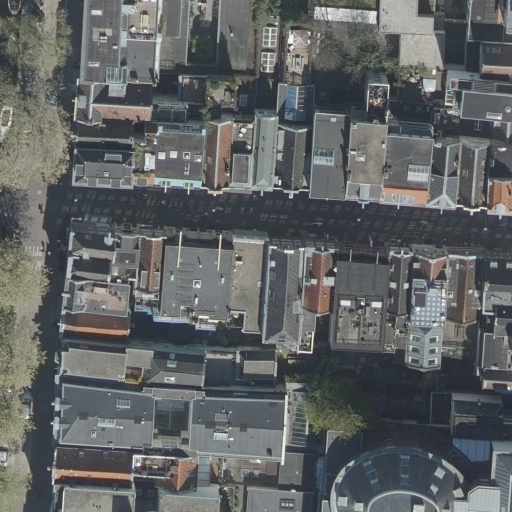}
  \subcaption*{\textbf{$Q$:} What color is the second small vehicle based on the left to right rule in this image?\\ \textbf{$A_{GT}$:} grey\\ \textbf{$A_P$:} grey}
  \end{subfigure}
  \begin{subfigure}[t]{0.23\linewidth}
  \includegraphics[width=0.95\linewidth]{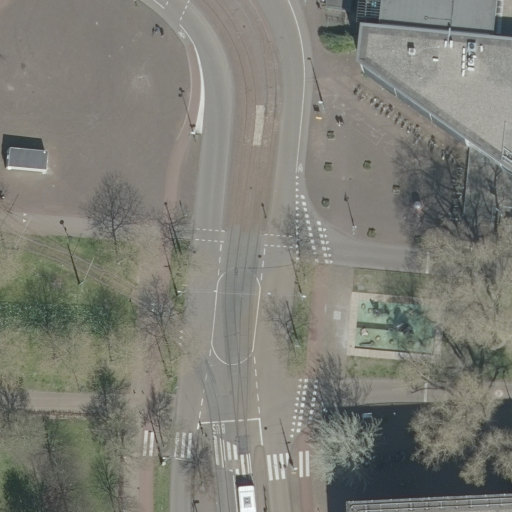}
  \subcaption*{\textbf{$Q$:} What kind of size is the first fitness field based on the left to right rule in this image?\\ \textbf{$A_{GT}$:} large scale\\ \textbf{$A_P$:} large scale}
  \end{subfigure}
  
  \begin{subfigure}[t]{0.23\linewidth}
  \includegraphics[width=0.95\linewidth]{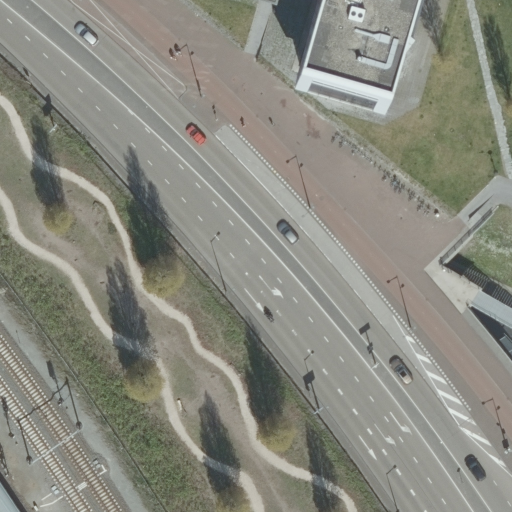}
  \subcaption*{\textbf{$Q$:} What kind of shape is the track in this image?\\ \textbf{$A_{GT}$:} linear\\ \textbf{$A_P$:} linear}
  \end{subfigure}
  \begin{subfigure}[t]{0.23\linewidth}
  \includegraphics[width=0.95\linewidth]{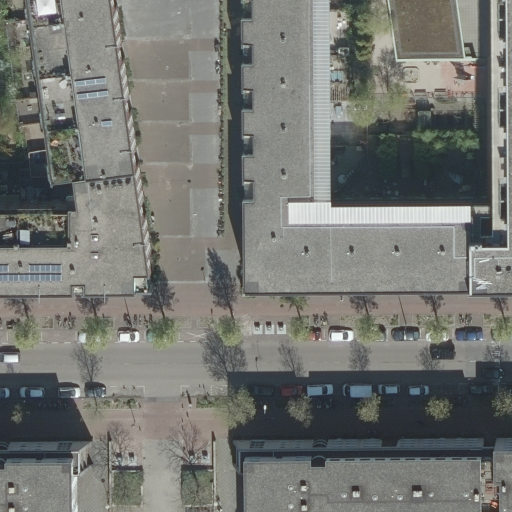}
  \subcaption*{\textbf{$Q$:} What is the main scene of this image?\\ \textbf{$A_{GT}$:} high density urban area\\ \textbf{$A_P$:} high density urban area}
  \end{subfigure}
  \begin{subfigure}[t]{0.23\linewidth}
  \includegraphics[width=0.95\linewidth]{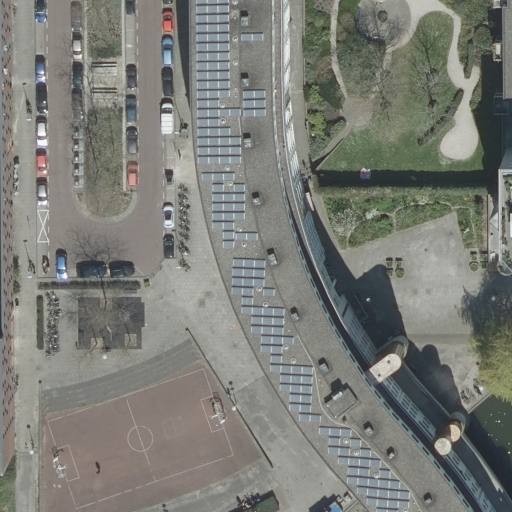}
  \subcaption*{\textbf{$Q$:} Does a building exist in this image?\\ \textbf{$A_{GT}$:} yes\\ \textbf{$A_P$:} yes}
  \end{subfigure}
  \begin{subfigure}[t]{0.23\linewidth}
  \includegraphics[width=0.95\linewidth]{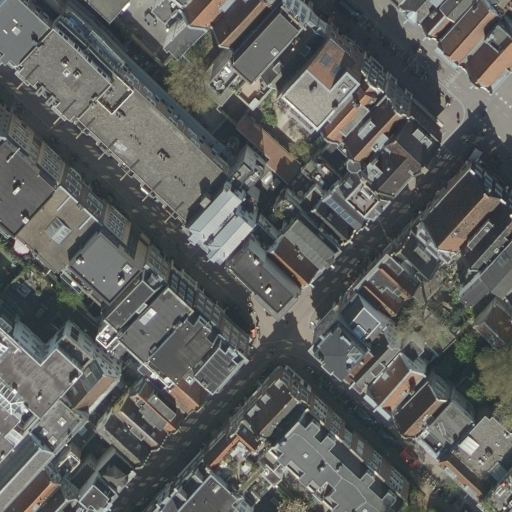}
  \subcaption*{\textbf{$Q$:} Where is the plane in this image?\\ \textbf{$A_{GT}$:} not found\\\textbf{$A_P$:} not found}
  \end{subfigure}
  
  \begin{subfigure}[t]{0.23\linewidth}
  \includegraphics[width=0.95\linewidth]{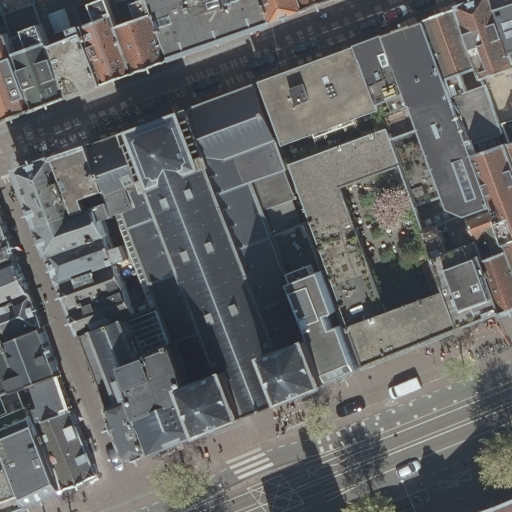}
  \subcaption*{\textbf{$Q$:} Are there less helipads than small vehicles in this image?\\ \textbf{$A_{GT}$:} yes\\ \textbf{$A_P$:} yes}
  \end{subfigure}
  \begin{subfigure}[t]{0.23\linewidth}
  \includegraphics[width=0.95\linewidth]{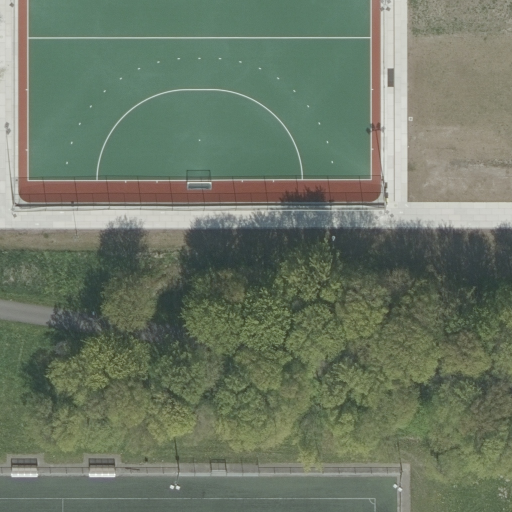}
  \subcaption*{\textbf{$Q$:} How can people improve their strength in this image?\\ \textbf{$A_{GT}$:} playing soccer\\ \textbf{$A_P$:} playing soccer}
  \end{subfigure}
  \begin{subfigure}[t]{0.23\linewidth}
  \includegraphics[width=0.95\linewidth]{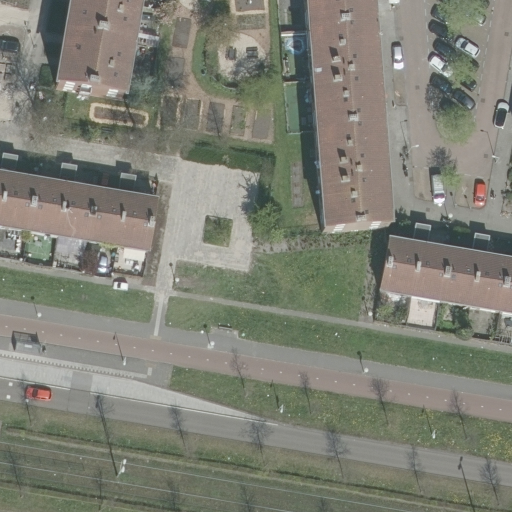}
  \subcaption*{\textbf{$Q$:} How many small vehicles are visible in this image?\\ \textbf{$A_{GT}$:} 14\\ \textbf{$A_P$:} 14}
  \end{subfigure}
  \begin{subfigure}[t]{0.23\linewidth}
  \includegraphics[width=0.95\linewidth]{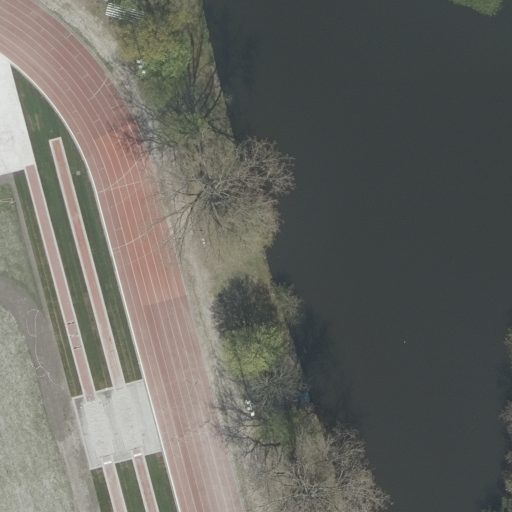}
  \subcaption*{\textbf{$Q$:} How big are the ground track fields in this image?\\ \textbf{$A_{GT}$:} more than 1000$m^2$\\ \textbf{$A_P$:} more than 1000$m^2$}
  \end{subfigure}
  \vspace{-2mm}
  \caption{Visual examples of predictions based on our method on HRVQA. \textbf{$A_{GT}$} means the ground truth while \textbf{$A_P$} means our prediction.
  }
  \vspace{-5mm}
  \label{fig:additional results}
\end{figure*}

Fig. \ref{fig:additional results} shows the predictions for different types of questions, and our method is capable of generating correct answers under various challenging scenes.

\begin{figure*}[ht]
  \centering
  \begin{subfigure}[t]{0.2\linewidth}
  \includegraphics[width=1\linewidth]{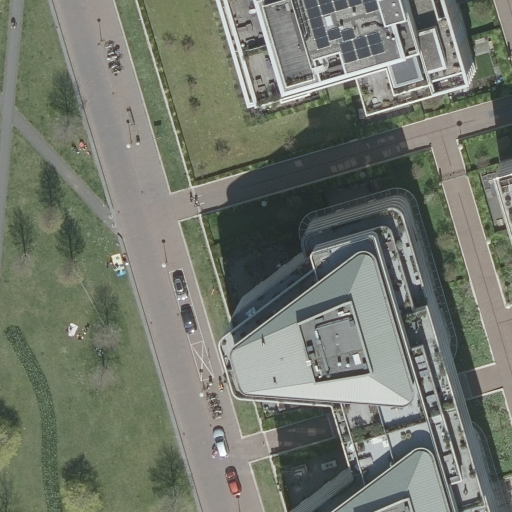}
  \subcaption*{\textbf{$Q$:} What is the color of the second small vehicle based on the left to right rule in this image?\\ \textbf{$A_{GT}$:} grey\\ \textbf{$A_P$:} blue}
  \end{subfigure}
  \hspace{2mm}
  \begin{subfigure}[t]{0.2\linewidth}
  \includegraphics[width=1\linewidth]{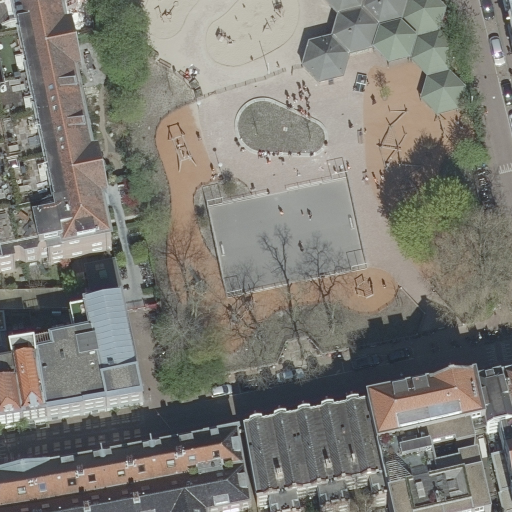}
  \subcaption*{\textbf{$Q$:} What sport can people do in this image?\\ \textbf{$A_{GT}$:} fitness or soccer\\ \textbf{$A_P$:} soccer}
  \end{subfigure}
  \hspace{2mm}
  \begin{subfigure}[t]{0.2\linewidth}
  \includegraphics[width=1\linewidth]{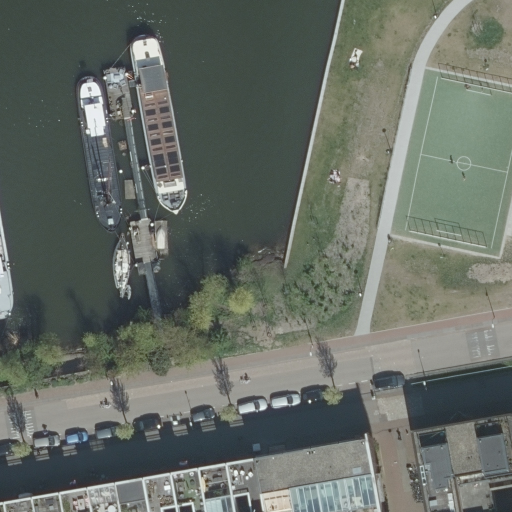}
  \subcaption*{\textbf{$Q$:} How many small vehicles can you see in this image?\\ \textbf{$A_{GT}$:} 10\\ \textbf{$A_P$:} 8}
  \end{subfigure}
  \hspace{2mm}
  \begin{subfigure}[t]{0.2\linewidth}
  \includegraphics[width=1\linewidth]{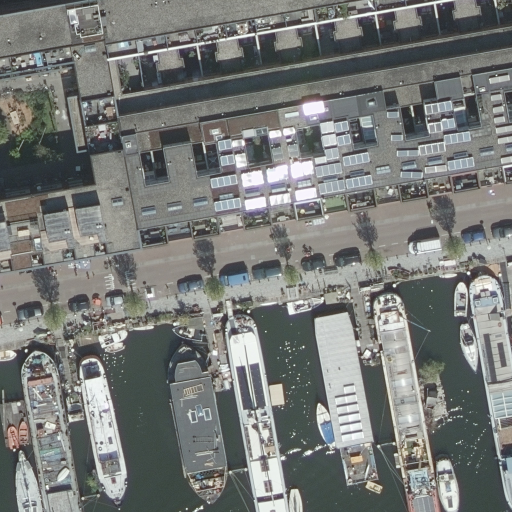}
  \subcaption*{\textbf{$Q$:} How many grey small vehicles can you see in this image?\\ \textbf{$A_{GT}$:} 6\\ \textbf{$A_P$:} 2}
  \end{subfigure}
  
  \begin{subfigure}[t]{0.2\linewidth}
  \includegraphics[width=1\linewidth]{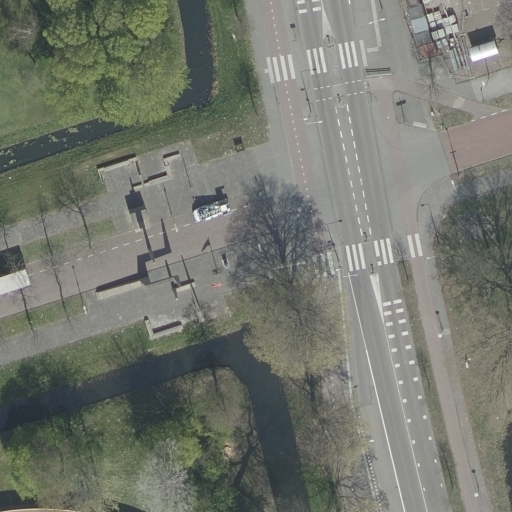}
  \subcaption*{\textbf{$Q$:} Does a building exist in this image?\\ \textbf{$A_{GT}$:} no\\ \textbf{$A_P$:} yes}
  \end{subfigure}
  \hspace{2mm}
  \begin{subfigure}[t]{0.2\linewidth}
  \includegraphics[width=1\linewidth]{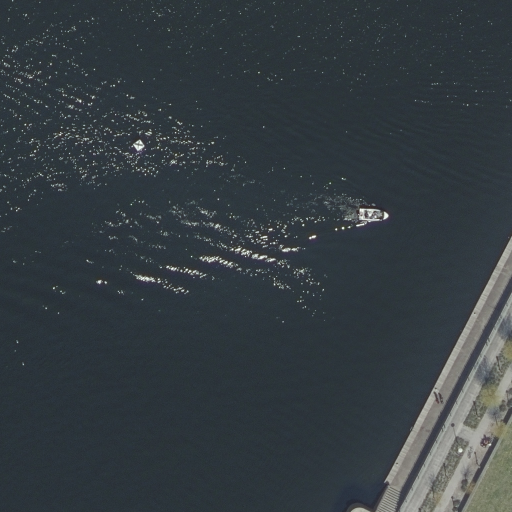}
  \subcaption*{\textbf{$Q$:} What kind of transportation can people use in this image?\\ \textbf{$A_{GT}$:} by ship\\\textbf{$A_P$:} on foot}
  \end{subfigure}
  \hspace{2mm}
  \begin{subfigure}[t]{0.2\linewidth}
  \includegraphics[width=1\linewidth]{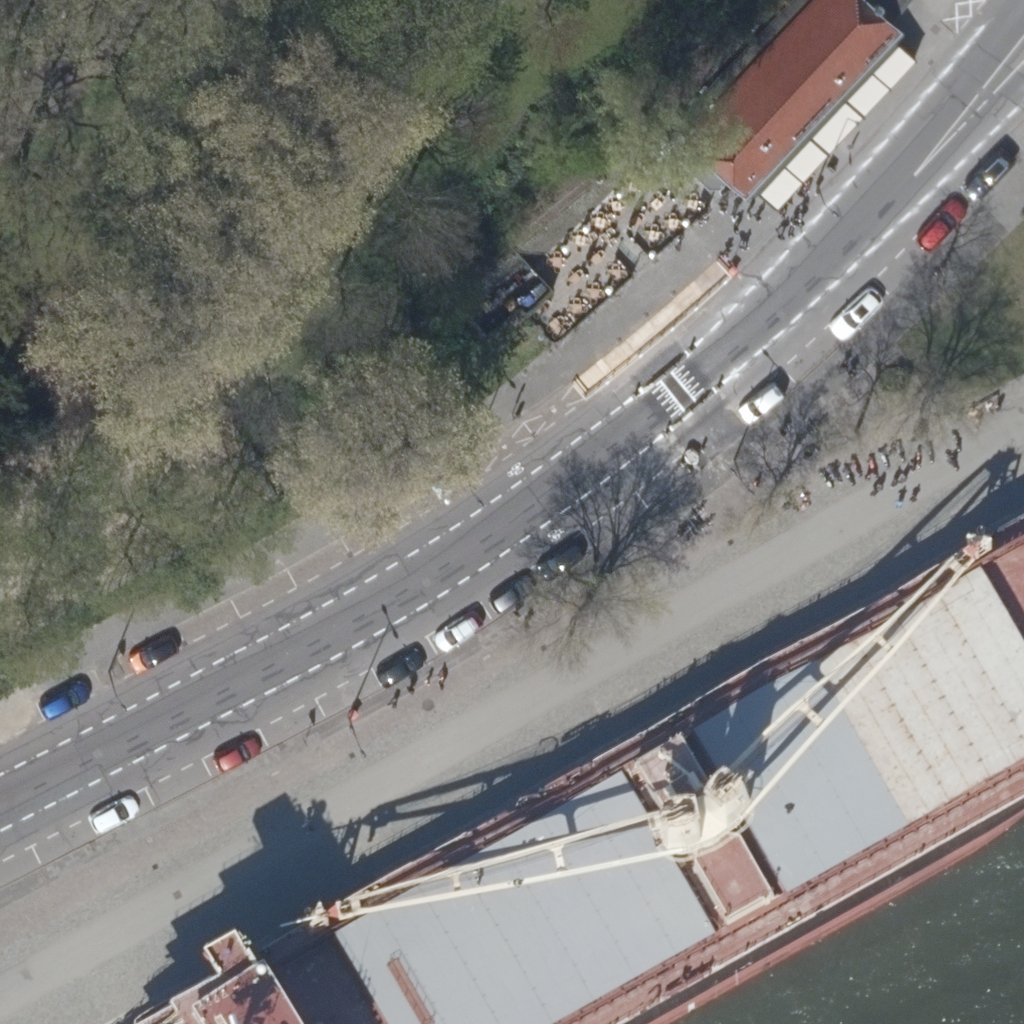}
  \subcaption*{\textbf{$Q$:} What color is the fourth small vehicle based on the left to right rule is this image?\\ \textbf{$A_{GT}$:} red\\ \textbf{$A_P$:} grey}
  \end{subfigure}
  \hspace{2mm}
  \begin{subfigure}[t]{0.2\linewidth}
  \includegraphics[width=1\linewidth]{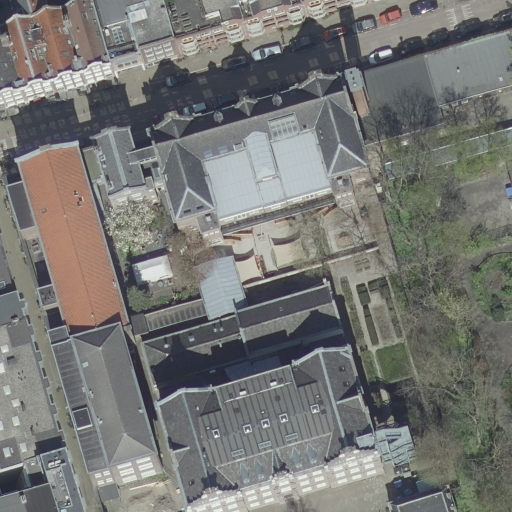}
  \subcaption*{\textbf{$Q$:} What is the main scene of this image?\\ \textbf{$A_{GT}$:} high density urban area\\ \textbf{$A_P$:} rural area}
  \end{subfigure}
  \vspace{-2mm}
  \caption{Random examples of failure cases with our method on HRVQA. \textbf{$A_{GT}$} means the ground truth while \textbf{$A_P$} means our prediction.}
  \vspace{-5mm}
  \label{fig:failure results}
\end{figure*}

Although our proposed method performs better than other models, some failure cases still need to be discussed to reveal points for further improvements. Fig. \ref{fig:failure results} shows some typical wrong predictions generated by our method. The questions about tiny objects from high-resolution aerial images are more difficult to answer compared to the ones from natural images, which leads the incorrect answers to some questions about specific attributes. Besides, a more easily understood descriptions on the target should be explored as the order of the concepts seems to add more burdens on models without any semantic or instance-level supervisions.

To further evaluate the robustness of the proposed method on RSVQA, we trained and tested different VQA models on the high-resolution subset (RSVQA-HR \cite{lobry2020rsvqa}). Note that we did not conduct comparative experiments with Swap-m \cite{gupta2022swapmix} for RSVQA-HR \cite{lobry2020rsvqa} because we do not have the concept information for the swapping operations only based on the available released annotations. Table \ref{tab:results in rsvqa test1} shows the remaining results on RSVQA-HR Test-1. We perform better than other models in general with 85.67\% overall accuracy and 85.90\% average accuracy. These results demonstrate that the proposed method is effective for the aerial image VQA. 

\begin{table}[t]
  \caption{Results of type-wise accuracy on RSVQA\cite{lobry2020rsvqa} Test-1.}
  \centering
  \setlength\tabcolsep{2.75pt}
  \begin{tabular}{p{2cm}cccccc}
    \toprule
    Method & Pres & Count & Comp & Area & OA & AA \\
    & -ence &  & -arison &  &  &  \\
    \midrule
    SAN\cite{yang2016stackedsan}$\ddagger$    & 90.21 & 67.53 & 88.47 & 81.65 & 82.45 & 81.97 \\
    MUTAN\cite{ben2017mutanmethod10}$\ddagger$  & 91.57 & 68.71 & 90.51 & 82.47 & 83.90 & 83.32 \\
    MCAN\cite{yu2019mcan}$\ast$   & 91.61 & 68.58 & 91.11 & 90.47 & 85.27 & 85.44 \\
    RSVQA\cite{lobry2020rsvqa}$\dagger$ & 90.43 & 68.63 & 88.19 & 85.24 & 83.23 & 83.12 \\
    M-CA\cite{martins2020sparse}$\ast$   & 91.46 & 68.71 & \textbf{91.33} & 89.88 & 85.24 & 85.34 \\
    TRAR\cite{zhou2021trar}$\ast$   & 91.24 & 68.97 & 91.00 & \textbf{91.33} & 85.36 & 85.64 \\
    AOA\cite{aoa}$\ast$ & \textbf{91.94} & 68.68 & 91.12 & 91.01 & 85.46 & 85.69 \\
    FETH\cite{yuan2022easy}$\dagger$ & 91.39 & 69.06 & 89.75 & 85.92 & 84.16 & 83.97 \\
    GFT (ours)$\ast$   & 91.80 & \textbf{69.47} & 91.18 & 91.14 & \textbf{85.67} & \textbf{85.90} \\
    \bottomrule
  \end{tabular}
  \label{tab:results in rsvqa test1}
\end{table}

\section{Ethics Statement}
\label{ethics statement}
We strongly advocate using our proposed dataset HRVQA properly. To advance the development of VQA for aerial images in the future, we will release the dataset and the relevant baseline codes. As the spatial resolution of the aerial images is very high (8 cm GSD~\footnote{GSD (Ground Sampling Distance) means the distance between pixel centers measured on the ground.}), some personal or commercial information (e.g., private nursery, club location and factory distribution) could be further discovered by analyzing the visual context. Note that human face will not be recognizable in such resolution, and other personal details like vehicle plate or house number information will not be detected either.

The ethics committee in our university has reviewed the ethical aspects of the proposed HRVQA, and has already \textit{approved} the dataset with respect to the concerns on geo-ethics and society.
To prevent the abusive usage of our proposed dataset, anyone who will use our data or method should mark with $research\ study$ and obey the corresponding regulations of dataset usage.

\newpage

{\small
\bibliographystyle{ieee_fullname}
\bibliography{egbib}
}

\end{document}